\documentclass[journal]{IEEEtran}
\ifCLASSINFOpdf
\else
\fi
\usepackage{color,soul}
\usepackage{graphicx}
\usepackage{longtable}
\usepackage{array}
\usepackage{booktabs}
\usepackage{amsfonts}
\usepackage{amsmath}
\usepackage{caption}
\usepackage{subcaption}
\usepackage{tabularx}
\usepackage{bm}
\usepackage{amsthm}
\usepackage[normalem]{ulem}
\usepackage[dvipsnames]{xcolor}
\usepackage[ruled,vlined,linesnumbered,lined,boxed,commentsnumbered]{algorithm2e}

\theoremstyle{definition}

\newtheorem{definition}{Definition}
\newtheorem{theorem}{Theorem}

\newcounter{mingcomm}
\definecolor{m}{rgb}{0.7,0.87,0.05}

\newcommand{\change}{\textcolor{black}}
\newcommand{\changesecond}{\textcolor{black}}
\newcommand{\changethird}{\textcolor{black}}
\newcommand{\specialcell}[2][c]{%
  \begin{tabular}[#1]{@{}c@{}}#2\end{tabular}}
\begin{document}
%
\title{Towards Graph Self-Supervised Learning with Contrastive Adjusted Zooming}
%
%
%
%

\raggedbottom
\author{Yizhen Zheng,
        Ming Jin,
        Shirui Pan, 
        Yuan-Fang Li, 
        Hao Peng,
        Ming Li,
        Zhao Li
\thanks{S. Pan is  supported by an ARC Future Fellowship (No. FT210100097). M. Li acknowledges the support from the National Natural Science Foundation of
China (No. 62172370). H. Peng acknowledges the support from the S\&T Program of Hebei through grant 21340301D.}
\thanks{Y. Zheng, M. Jin, and Y.-F. Li are with Department of Data Science and AI, Faculty of IT, Monash University, Clayton, VIC 3800, Australia. Emails: \{yizhen.zheng, ming.jin, yuanfang.li\}@monash.edu}
\thanks{S. Pan is with School of Information and Communication Technology, Griffith University, Australia, SouthPort, QLD 4222, Australia. Emails:s.pan@griffith.edu.au (\textit{Corresponding Author: Shirui Pan}).}
\thanks{H. Peng is with Beijing Advanced Innovation Center for Big Data and Brain Computing, Beihang University, Beijing 100191, China. Email: penghao@buaa.edu.cn.}
\thanks{M. Li is with Key Laboratory of Intelligent Education Technology and Application of Zhejiang Province, Zhejiang Normal University, Jinhua 321004, China. Email: mingli@zjnu.edu.cn.}
 \thanks{Z. Li is with Alibaba Group, Hangzhou 311100, China. Email: lizhao.lz@alibaba-inc.com}
\thanks{Manuscript received May 31, 2021; revised xx xx, 202x.}
}

%
%

\markboth{Journal of \LaTeX\ Class Files,~Vol.~14, No.~8, March~2021}%
{Shell \MakeLowercase{\textit{et al.}}: Graph Contrastive Learning via Adjusted Zooming}
%



\maketitle

\begin{abstract}
 Graph representation learning (GRL) is critical for graph-structured data analysis. However, most of the existing graph neural networks (GNNs) heavily rely on labeling information, which is normally expensive to obtain in the real world.
 Although some existing works aim to effectively learn graph representations in an unsupervised manner, they suffer from certain limitations, such as the heavy reliance on monotone contrastiveness and limited scalability.
 To overcome the aforementioned problems, in light of the recent advancements in graph contrastive learning, we introduce a novel self-supervised graph representation learning algorithm via \underline{G}raph Contrastive Adjusted \underline{Zoom}ing, namely \textbf{G-Zoom}, to learn node representations by leveraging the proposed \textbf{\textit{adjusted zooming}} scheme. Specifically, this mechanism enables G-Zoom to explore and extract self-supervision signals from a graph from multiple scales:
\textit{micro} (i.e.,\ node-level), \textit{meso} (i.e.,\ neighborhood-level), and \textit{macro} (i.e.,\ subgraph-level).
 Firstly, we generate two augmented views of the input graph via two different graph augmentations. 
 Then, we establish three different contrastiveness on the above three scales progressively, from node, neighboring, to subgraph level, where we maximize the agreement between graph representations across scales. 
 While we can extract valuable clues from a given graph on the micro and macro perspectives, 
 the neighboring-level contrastiveness offers G-Zoom the capability of a customizable option based on our adjusted zooming scheme
 to manually choose an optimal viewpoint that lies between the micro and macro perspectives to better understand  the graph data. Additionally, to make our model scalable to large graphs, we employ  a parallel graph diffusion approach to decouple model training from the graph size. We have conducted extensive experiments on real-world datasets, and the results demonstrate that our proposed model outperforms state-of-the-art methods consistently.
 

\end{abstract}


%
\IEEEpeerreviewmaketitle

\section{Introduction}
\IEEEPARstart{G}{raph} representation learning (GRL) has become a pivotal strategy for analyzing semi-structured graph data in recent years. GRL aims to distill the high-dimensional structural and attributive information from graphs to generate low-dimensional embeddings for nodes or graphs. These learned embeddings can then be used in various downstream tasks such as node classification \cite{kipf2016semi, zhang2022trustworthy}, graph classification \cite{sun2019infograph} and link prediction~\cite{schutt2017schnet}. It has been applied to many real-world graph datasets from different domains \cite{xia2021graph} such as social networks \cite{kipf2016semi}, geoscience \cite{wan2021dual}, anomaly detection \cite{liu2021anomaly, yizhen2021hetergraph} and molecules \cite{dai2016discriminative}. 

Recently, existing successful GRL approaches are mainly powered by Graph Neural Networks (GNN) \cite{kipf2016semi, velivckovic2017graph, wu2019simplifying}, which learn low-dimensional node embeddings via iterative message passing to aggregate topological representations of neighbors. However, most of these methods adopt supervised learning and rely extensively on labeling information, which is both laborious and expensive to collect in the real world. To address this issue, some self-supervised GRL methods based on contrastive learning emerged, e.g., Deep Graph Infomax (DGI) \cite{velickovic2019deep}, Graph Contrastive Representation Learning (GRACE) \cite{zhu2020deep} and Graphical Mutual Information (GMI) \cite{peng2020graph}. The basic idea of these approaches is to set up pretext tasks without using 
labeling information to train a GNN encoder for node embeddings generation. While DGI  \cite{velickovic2019deep} focuses on maximizing the Mutual Information (MI) between 
node- and graph-level representations, GRACE \cite{zhu2020deep} and GMI \cite{peng2020graph} extend this MI maximization scheme to contrast node-level representations in two different graph views or enlarge the MI between the hidden representation of nodes and the raw node features of their one-hop neighbors.

Though these methods have achieved promising results, they adopt 
monotonous contrastive learning on graphs, which only inspects graph-structured data from a single fixed perspective. For example, DGI and MVGRL only conduct contrasting between patch and global representations, 
i.e., contrasting from a global viewpoint. 
Without considering the contrastiveness from a macro perspective, other methods such as GRACE, SubG-Con \cite{jiao2020sub}, and GMI only emphasize the local structure of target nodes by maximizing agreement between patch representations in two augmented views, or between 
nodes and representations of their neighbors. 
However, contrasting across different scales benefits the node representation learning by injecting richer contextual information into the discrimination \cite{liu2021graph}. 
While global contextual information can be extracted by discriminating the node-level embeddings with the graph-level representations, node- and neighboring-level contrastiveness emphasize encoding the local contextual information. 
Thus, 
the aforementioned contrastive learning schemes 
either neglect rich global information or localized information reflected by different viewpoints of graphs. 
As a result, these methods may fail to obtain high-quality node representations and suffer from poor performance on downstream tasks due to their monotonous contrastive schemes. 

\begin{figure}
\centering\includegraphics[scale = 0.3]{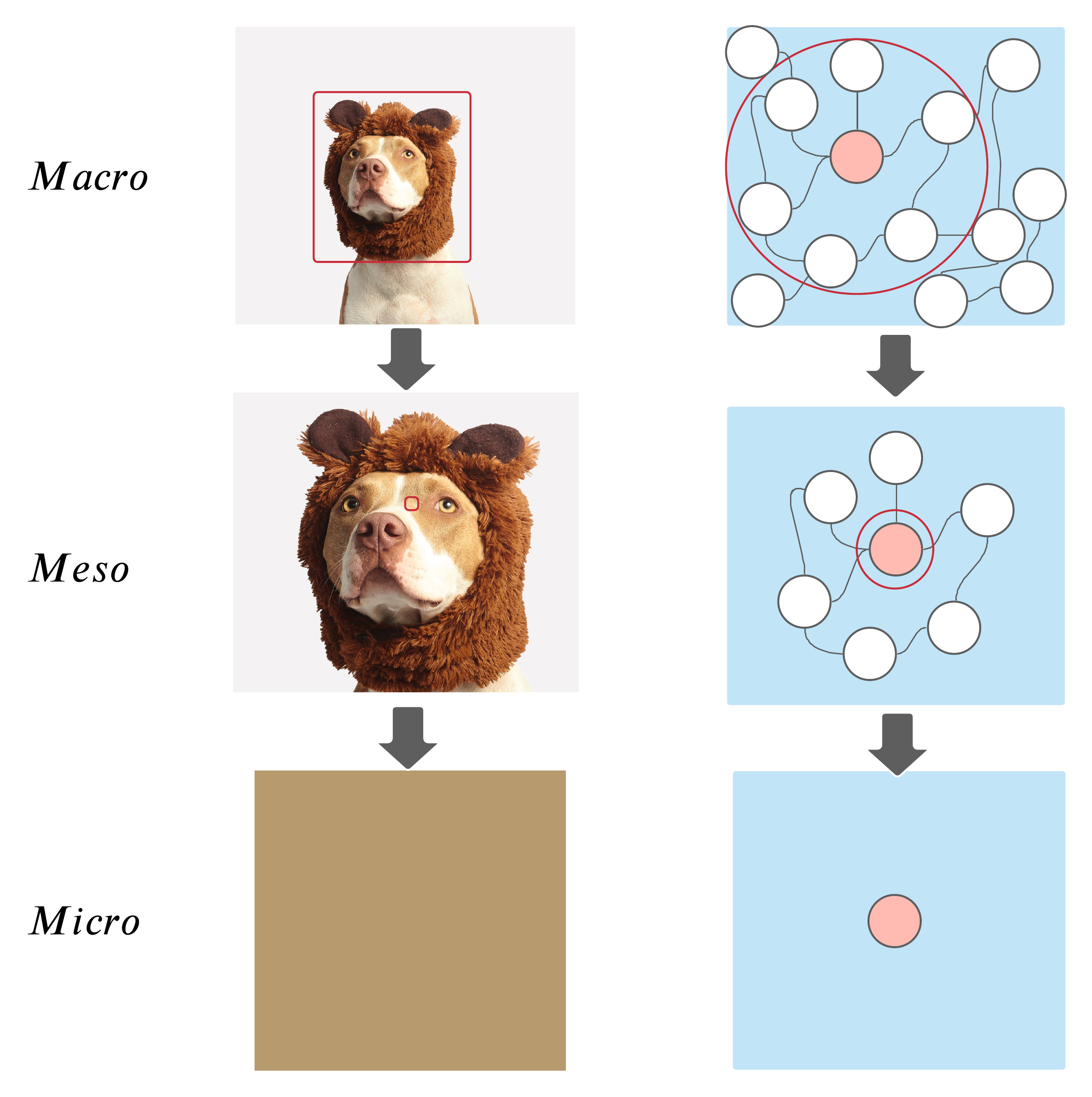}
\caption{\small An analogy between image and graph adjusted zooming. The red squared in the image and the circled selected area on the sample graph are chosen to be enlarged in the next level perspective. The red node represents the target node of the sample graph.}
\label{fig:1}
\end{figure}

In this paper, to overcome the aforementioned problems, we propose a novel graph self-supervised representation learning algorithm with a novel \underline{G}raph Contrastive Adjusted \underline{Zoom}ing mechanism, named G-Zoom. 
To present this scheme, 
we have made an analogous explanation between the image scaling and our proposed graph adjusted zooming, as illustrated in Figure \ref{fig:1}. 
Similar to the vision domain, 
graphs can also be inspected from 
three different 
perspectives: micro, meso, and macro.
From the macro perspective, the broadest zooming scale in our method allows us to examine the entire image or graph.  
In contrast, from the micro viewpoint, we can inspect the finest detail of a selected area of the data, i.e., a pixel for an image or a node for a graph. Standing between the micro and the macro perspective, inspection from the meso perspective allows us to choose a specific area, i.e., (a customized degree of zooming level) 
to exploit the valuable information embodied in this viewpoint. For instance, for the dog image presented in Figure \ref{fig:1}, 
we can select a specific part (e.g., head of the dog) and then zooming to an extent where we can see the object clearly (e.g., dog head). 
Similarly, in a graph, we can choose a target node and then decide the scope of its contextual information we want to examine. 

In general, it is straightforward to find that different perspectives offer different scenes and contextual information. 
While from the macro perspective, we zoom out 
observing the entire picture and ignore the detailed information, 
from the micro 
perspective, we concentrate on the microscopic element of the data regardless of the broader view. 
Different 
from other two perspectives, 
the meso perspective offers a more flexible viewpoint, 
enabling us to inspect a specific area with the desired scale. 
\change{This perspective forces the model to explore the fine-grained information embodied in this scale, which can easily be neglected from the macro-level viewpoint. As such, it complements the macro perspective with overlooked detailed information in the semi-global context.} Therefore, it is necessary to include various perspectives on multiple scales when conducting graph contrastive learning. 

In G-Zoom, we propose to formulate three different contrastive paths on various scales progressively: node-, neighboring-, and subgraph-level. \change{Inspired by GRACE and MVGRL, we establish node- and subgraph-level contrastiveness to facilitate discrimination from the micro and macro perspective. Then, with our proposed adjusted zooming scheme, we design the neighboring-level discrimination path, which provides a tailored view of contrastiveness.}
 Specifically, two graph views are firstly sampled with different graph augmentations (e.g., graph sampling and diffusion). 
Then, we input two augmented views into a GNN encoder to get node representations in sampled graphs. At this stage, we can readout the subgraph-level representations via the graph pooling and obtain neighboring-level representations by aggregating the top-$k$ neighborhood representations of selected central nodes. 
Finally, we formulate our training loss 
by summing up the contrastive losses on three different contrasting levels to train the graph encoder for the node embedding calculation. Compared with prior arts, G-Zoom not only facilitates multi-scale contrasting to provide learnable clues from multiple perspectives for model training but also can be extended to handle large-sized graphs. 
Experimental results on a variety of datasets have shown the superiority of G-Zoom compared with several state-of-the-art baselines. The main contributions of our work are listed as below: 
\begin{itemize}
    \item 
    We propose a novel algorithm, namely G-Zoom, to learn graph representation in an unsupervised manner. To the best of our knowledge, this is the first attempt to conduct graph contrastive learning on multiple scales that crossed three different graph topological levels.
    \item We propose the graph adjusted zooming mechanism, which aims to facilitate the multi-scale graph contrastive learning from various perspectives to enrich the self-supervision signals and overcome the limitations of existing works.
    \item We conduct extensive experiments on real-world datasets to validate \changesecond{the effectiveness of our proposed approach}. The results show that our model consistently outperforms state-of-the-art baselines.
\end{itemize}
%
%

\section{Related Work}
In this section, we review related works in three areas: graph neural networks, unsupervised graph representation learning, and contrastive self-supervised \changethird{l}earning.

\subsection{Graph Neural Networks}
Under the umbrella of deep neural network \cite{lecun2015deep}, Graph Neural Networks (GNNs) learn node embeddings by utilizing both attributive and topological information of non-Euclidean graph-structured data. GNN is firstly introduced in \cite{bruna2013spectral}, which proposes a spectral-based method extending convolution networks to graphs. Then, a series of subsequent spectral-based convolution GNNs has been introduced \cite{defferrard2016convolutional,henaff2015deep}, which adopts filters in light of graph signal processing \cite{shuman2013emerging}. Notably, bridging the gap between spectral-based and spatial-based GNN methods, GCN \cite{kipf2016semi} simplifies graph convolutions by approximating spectral-based graph convolutions with stacked first-order Chebyshev polynomial filters. After that, spatial-based approaches grow promptly because of their advantages of efficiency and general applicability,  
e.g., GAT \cite{velivckovic2017graph} incorporates the attention mechanism \cite{vaswani2017attention} to consider the difference of importance among node neighbors rather than simply averaging neighbors information. SGC \cite{wu2019simplifying} reduces the excess complexity of GCN via removing non-linearity and collapsing weight matrices between graph convolution layers. Except for SGC, many studies are aiming to improve GNN from different perspectives such as scalability extension \cite{bojchevski2020scaling, zeng2019graphsaint}, and receptive field enlargement \cite{klicpera2019diffusion, tang2020cgd, liao2019lanczosnet,wu2021learning}. Existing GNN approaches have been successfully applied in various domains, e.g., bankruptcy prediction \cite{yizhen2021hetergraph}, knowledge graph \cite{ji2021survey, zhang2020relational} and traffic prediction \cite{zhao2019t}.

Though there are many off-the-shelf GNNs available, many of them rely extensively on labeling information, which are unable to deal with unlabeled graphs. However, unlabeled graphs are pervasive in the real world. To tackle this problem, we proposed G-Zoom, which utilizes an adjusted-zooming-based contrastive learning scheme to generate effective node embeddings without the guidance of node labels. In G-Zoom, we select GCN as the kernel of our GNN encoder component. Except for GCN, the encoder component of G-Zoom can be underpinned by any aforementioned GNNs.

\subsection{Unsupervised Graph Representation Learning}
Unsupervised graph representation learning aims to learn node embeddings on unlabeled graphs, in which node labels are unavailable. Mainly, there are two traditional ways for unsupervised algorithms to exploit the underlying topological information. One way is adopting an auto-encoder, which consists of an encoder for embedding a graph into latent representation and a decoder for reconstructing the topology of the graph \cite{kipf2016variational, pan2019learning}. These autoencoder-based methods heavily rely on the assumption that neighboring nodes have similar representations. The effectiveness of objectives based on this assumption is questionable \cite{jiao2020sub}. Another way conducts model optimization by utilizing random-walk-based objectives \cite{grover2016node2vec,perozzi2014deepwalk,hamilton2017inductive}. However, these methods are criticized because of having one or multiple issues including overwhelmingly emphasizing structural information embodied in graph proximity, limited scalability, and the inability of learning attributive information \cite{zhu2020deep, hassani2020contrastive}.

\subsection{Contrastive Self-supervised Learning}
Recently, there are some successful applications of using self-supervised contrastive learning approaches for image processing \cite{chen2020simple, hjelm2018learning, caron2020unsupervised, he2020momentum, grill2020bootstrap} and have achieved promising results. With the help of defined annotation-free pretext tasks, these methods train models via contrasting positive and negative instance pairs to calculate the contrastiveness loss. Some recent works also attempt to adapt contrastive methods to the graph domain. Inspired by Deep Infomax \cite{caron2020unsupervised}, Deep Graph Infomax (DGI)  \cite{velickovic2019deep} pioneers contrastive learning for graphs. Specifically, DGI considers node and graph-level representations obtained with a readout function as positive pairs and utilizes graph corruption techniques to conduct negative sampling. On the basis of DGI, Multi-view graph representation learning (MVGRL) \cite{hassani2020contrastive} consolidates \changesecond{the patch-summary contrastive learning approach of DGI} by integrating graph augmentation techniques including graph sampling and graph diffusion to generate multiple augmented graph views. Graph Contrastive Representation Learning (GRACE) \cite{zhu2020deep} focuses on node-level contrasting between two augmented views, which are built through attributive and structural modification on the input graph. Graph mutual information (GMI) and Sub-graph Contrast (SubG-Con) aim to encode localized signals by contrasting the central node representation with the representations of their close neighbors.

Although these graph self-supervised approaches fulfill comparable performance to supervised methods, they are still suffering from issues such as monotonous contrastive learning schemes and limited scalability. To be specific, DGI, MVGRL, GRACE, and GMI fail to handle large-scale graphs, and all aforementioned methods are incapable of encoding information of a graph thoroughly and can only extract information from a single perspective, i.e., micro, macro, or meso. To address these issues, we proposed an adjusted-zooming contrastive learning mechanism for comprehensive multi-scale contrasting in G-Zoom. Also, G-Zoom can be extended to handle large-scale graphs. The detailed explanation of this extension is shown in Subsection \ref{section:IV-E}. Extensive experiments in the following sections have shown \changesecond{the superiority of G-Zoom in graph representation learning.}

%
%


\section{Problem Formulation}
In this section, we introduce the problem of unsupervised node representation learning. The notations used in this paper are summarized in Table \ref{table:notation}. In this paper, we use bold uppercase letters (e.g., \textbf{X}) and bold lowercase letters (e.g., \textbf{x}) to represent matrices and vectors, respectively.

Let $\mathcal{G}=(\textbf{X}, \textbf{A})$ denotes a graph with a node feature matrix $\textbf{X} \in \mathbb{R}^{n \times d}$ and an adjacency matrix $\textbf{A} \in \mathbb{R}^{n \times n}$, 
in which \changesecond{the value at the i-th row in the j-th column of \textbf{A} is} 1 if node $v_i$ and $v_j$ are connected. 
Here $n$ and $d$ denotes the number and dimension of nodes in $\mathcal{G}$, respectively. 
In G-Zoom, we leverage a GNN as the backbone encoder to transform the original high-dimensional features into a low-dimensional representation, which has been defined below:

\begin{definition}[Graph Neural Networks]
Given an attributed graph $\mathcal{G}=(\textbf{X}, \textbf{A})$, a typical graph neural network mainly consists of two components: \changesecond{message aggregation} and transformation:
\begin{align}
\mathbf{m}_{\mathcal{G}}^{(l)}[v_t,:] &= aggregate^{(l)}(\mathbf{H}_{\mathcal{G}}^{(l-1)}[v_k,:]: v_k \in \mathcal{N}(v_t)) \nonumber, \\
\mathbf{H}_{\mathcal{G}}^{(l)}[v_t,:] &= transform^{(l)}(\mathbf{H}_{\mathcal{G}}^{(l-1)}[v_t,:], \mathbf{m}_{\mathcal{G}}^{(l)}[v_t,:]) \label{eq: gnn},
\end{align}
where $aggregate$ and $transform$ are message aggregation and transformation functions. $v_t$ represents the target node whose embedding to be learned. $l$ in above formulas denotes the $l$-th layer. \changesecond{$\mathcal{N}(v_t)$ is the set of neighboring nodes for $v_t$. $\mathbf{m}_{\mathcal{G}}^{(l)}$ is the aggregated message encapsulating information from neighboring nodes, whose dimension size is $n \times d$ when $l = 1$ and $n \times d'$ when $l > 1$.}
In general, a graph neural network generates the representation of a node by combining representation of its own and its neighbors.
\end{definition}
The main focus of this work is to conduct the unsupervised node representation learning on graphs, which is defined as:

\begin{definition}[Unsupervised node representation learning]
\changesecond{Provided an attributed graph $\mathcal{G}=(\textbf{X} \in \mathbb{R}^{n \times d}, \textbf{A} \in \mathbb{R}^{n \times n})$, where $n$ is number of nodes and $d$ is the feature dimension, our aim is to train an effective GNN $g_{\theta}$, which learns node embeddings $\textbf{H} \in \mathbb{R}^{n \times d'}$ from $\mathcal{G}=(\textbf{X}, \textbf{A})$, where $d'$ is the hidden and the output dimension size}, without relying on any labeling information.
\end{definition}

\changesecond{Finally, by passing the output embeddings $\textbf{H}$ to an output classifier}, our model is able to handle various downstream tasks, such as node classification.


    \begin{table}[t]
	\small
	\centering
	\caption{Summary of the primary notations. \change{Here, notations in bold uppercase letters (e.g.,\textbf{X}) represent matrices, while bold lowercase letters (e.g.,\textbf{h}) mean vectors.}} 
	\begin{tabular}{ p{62 pt}<{\centering} | p{159 pt}}  
		\toprule[1.0pt]
		Symbols & Description  \\
		\cmidrule{1-2}
		$\mathcal{G}=(\mathbf{X}, \mathbf{A})$ & A graph with feature matrix \\
		\changesecond{$\mathbf{X} \in \mathbb{R}^{n \times d}$} & The feature matrix of $\mathcal{G}$ \\
		$\mathbf{A} \in \mathbb{R}^{n \times n}$ & The adjacency matrix of $\mathcal{G}$ \\
		$n$ & Number of nodes in $\mathcal{G}$\vspace{0.5mm}\\
        $d$ & Number of dimensions of $\mathbf{X}$ \vspace{0.5mm}\\
        $d'$ & Number of dimensions of latent representations \vspace{0.5mm}\\
		\cmidrule{1-2}
        $\tilde{\mathcal{G}} = (\textbf{X}, \tilde{\textbf{A}})$ &  The diffused version of the input graph $\mathcal{G}$ \vspace{0.5mm}\\
        $\hat{\mathcal{G}} = (\hat{\textbf{X}}, \hat{\textbf{A}})$ &  A sampled graph derived from $\mathcal{G}$ \vspace{0.5mm}\\
        $B$ &  Number of nodes in a batch \vspace{0.5mm}\\
        $P$ &  Number of nodes in a $\hat{\mathcal{G}}$ \vspace{0.5mm}\\
        $v_t$ &  The target node \vspace{0.5mm}\\
        $k$ &  \changethird{Number of top-ranked neighbors for $v_t$}  \vspace{0.5mm}\\
		\cmidrule{1-2}
        $\hat{\mathcal{G}_1}, \hat{\mathcal{G}_2}$ &  The augmented view of $\mathcal{G}$ \vspace{0.5mm}\\
        $g_{\theta}$ & The GNN encoder used in our model\\
        $\textbf{H}_1, \textbf{H}_2 \in \mathbb{R}^{n \times d'}$ & The latent representation of $\hat{\mathcal{G}_1}$ and $\hat{\mathcal{G}_2}$ \changethird{encoded via} $g_{\theta}$ \vspace{0.5mm}\\ 
        $\tilde{\textbf{H}} \in \mathbb{R}^{n \times d'}$ & Corrupted representation for $\mathcal{G}$ \vspace{0.5mm}\\ 
        $\vec{\textbf{H}} \in \mathbb{R}^{n \times d'} $ & Final embedding matrix for $\mathcal{G}$ \vspace{0.5mm}\\ 
        \changethird{$\textbf{h}^t_1, \textbf{h}^t_2 \in \mathbb{R}^{d'}$} &  \changesecond{Node-level representation of $v_t$ in $\textbf{H}_1, \textbf{H}_2$}\vspace{0.5mm}\\
        \changethird{$\textbf{s}_1, \textbf{s}_2 \in \mathbb{R}^{d'}$} &  Subgraph-level representation for $\hat{\mathcal{G}_1}$,$\hat{\mathcal{G}_2}$\vspace{0.5mm}\\
        \changethird{$\textbf{n}_1^t, \textbf{n}_2^t \in \mathbb{R}^{d'}$} &  \changesecond{Neighboring-level representation of $v_t$} in $\textbf{H}_1, \textbf{H}_2$ \vspace{0.5mm}\\
        $\mathcal{D}$ &  Discriminator for mutual information evaluation\vspace{0.5mm}\\
        \changesecond{$\textbf{W}^{l}$} & The $l$-th layer trainable weight matrix in $g_{\theta}$. \changesecond{If $l = 1$, $\textbf{W}^l \in \mathbb{R}^{d \times d'}$, while $l > 1$,  $\textbf{W}^l \in \mathbb{R}^{d' \times d'}$.}\\
        $\textbf{W}_D \in \mathbb{R}^{d' \times d'}$ & The trainable weight matrix in $\mathcal{D}$\\
		\bottomrule[1.0pt]
	\end{tabular}
	\label{table:notation}
\end{table}

\section{methodology}

\begin{figure*}
\centering
\includegraphics[scale = 0.16]{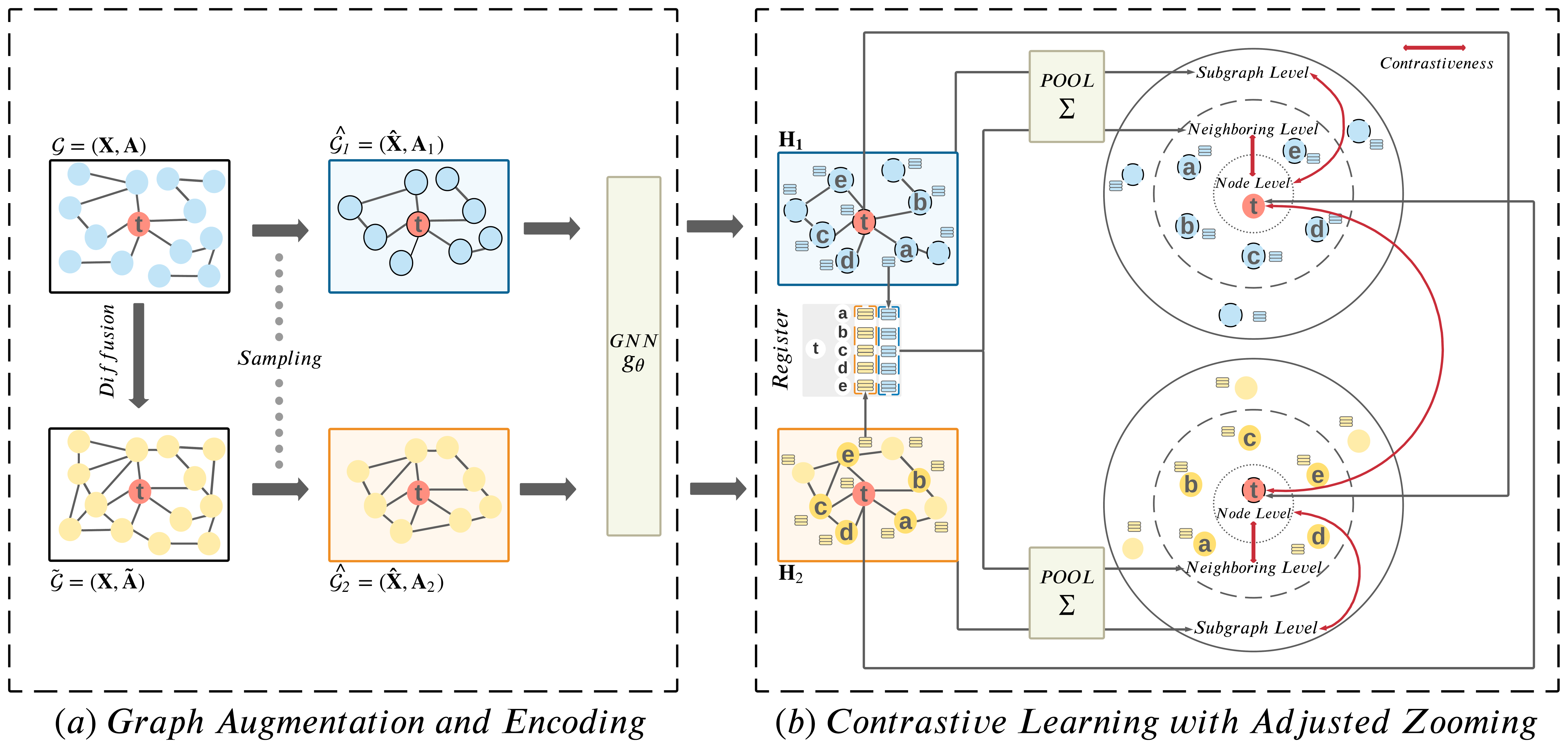}
\caption{\small The overall architecture of G-Zoom, which consists of two main components: Graph Augmentation and Encoding, and Contrastive Learning with Adjusted Zooming. Here we consider the red-colored node $v_t$ as the target node. In the first component, we generate two augmented views (i.e., $\hat{\mathcal{G}}_1$ and $\hat{\mathcal{G}}_2$) and input these two views to the GNN encoder $g_{\theta}$ to obtain representations for these two views $\textbf{H}_1$ and $\textbf{H}_2$. Then, in the second component which we utilizes a neighborhood register to attain \changesecond{the top-$k$ strongly correlated neighbors of $v_t$}, which are  $v_a$, $v_b$, $v_c$, $v_d$ and $v_e$. After that, we use a pooling layer to aggregate representations of these neighbors to get the neighboring-level representation for $v_t$. Also, we can obtain the subgraph-level embedding by aggregating all node embeddings in views via the same pooling layer. Finally, by establishing the contrastive paths between \changesecond{the representation of node $v_t$} in one view and the other two-level representations in the other view, we can effectively encode valuable information for learning \changesecond{the representation of $v_t$.}}
\label{fig:2}
\end{figure*}

In this section, we present the proposed G-Zoom algorithm. 
As shown in Figure \ref{fig:2}, our model consists of two main components, which are \textit{graph augmentation and encoding}, and \textit{contrastive learning with adjusted zooming}. 
Firstly, we apply graph augmentations including graph diffusion and sampling to generate two augmented views based on the input graph. While the first augmented view $\hat{\mathcal{G}_1}$ is created by applying graph sampling on the input graph $\mathcal{G}$ directly, the second congruent view $\hat{\mathcal{G}_2}$ is sampled from the diffused graph $\tilde{\mathcal{G}}$. Then, by feeding these two views to the GNN encoder $g_{\theta}$, 
we map node features to the latent space, 
denoted as $\textbf{H}_1$ and $\textbf{H}_2$ for $\hat{\mathcal{G}_1}$ and $\hat{\mathcal{G}_2}$, respectively. 
This process is shown in the leftmost part in Figure \ref{fig:2}. 
After this, we set up different contrastive paths on multiple scales via the proposed adjusted zooming scheme to facilitate the extraction of self-supervision signals for our model training. 
Given a target node $v_t$, we can get its top-$k$ correlated 
neighbors by 
designing 
a neighborhood register\changesecond{, which is defined in Definition \ref{def:register} \changethird{(Defined in Subsection \ref{meso(neighboring-level) contrastiveness})}. Then, we aggregate their 
embeddings through a pooling
layer 
to obtain \changesecond{the neighboring-level representation of $v_t$, }$\textbf{n}_1^t$ and $\textbf{n}_2^t$ for the two views.}
To obtain the subgraph-level representation $\textbf{s}_1$ and $\textbf{s}_2$ for $\hat{\mathcal{G}_1}$ and $\hat{\mathcal{G}_2}$, 
we readout the embeddings of nodes in $\textbf{H}_1$ and $\textbf{H}_2$ via the same pooling layer. 
At this stage, we conduct 
contrastive learning through the adjusted zooming scheme on 
three perspectives, namely micro, meso, and macro by contrasting among node-, neighboring-, and subgraph-level representations 
of the target node $v_t$ in two augmented views. 

The following section will discuss the two main segments of our framework in detail (Subsection \ref{section:IV-A} to \ref{section:IV-B}) and the model training and algorithm of G-Zoom (Subsection \ref{section:IV-C}). 

\subsection{Graph Augmentations and Encoding}
\label{section:IV-A}
Recently, there are some successful attempts on 
self-supervised visual representation learning, which allow encoders to learn effective representations by contrasting different augmented views of images \cite{hjelm2018learning}. However, we can not directly employ standard image augmentations 
such as image cropping, rotation, and distortion, to graphs due to their 
complex properties, e.g., no spatial locality and fixed node ordering. To address this issue, there are mainly two types of augmentations on graphs: topological and attributive. 
While the first type of 
strategy augments 
\changesecond{the topological space of a graph} via the graph sampling (GS), edge dropout (ED), or graph diffusion (GD), the latter one conducts the augmentation operation on attributive properties of nodes
, such as node feature dropout (NFD), which masks a fraction of node feature dimensions with zeros. 

In our model, we mainly 
apply topological graph augmentations 
to establish different graph views. 
The reason is not only because there are many unattributed graphs in the real world, but also through our experiments (i.e., Subsection \ref{section:V-E}), we find that attributive augmentations bring limited performance gain to G-Zoom. 
Specifically, we employ GS to generate a congruent view $\hat{\mathcal{G}_1}$ and use the combination of GD and GS to create an incongruent view $\hat{\mathcal{G}_2}$ of the input graph. \change{Please note} that all the nodes in  $\hat{\mathcal{G}_1}$ are the same as $\hat{\mathcal{G}_2}$ since they share the same sub-sampling scheme. 
Previous works on graph contrastive learning, such as MERIT~\cite{jin2021merit} and MVGRL~\cite{hassani2020contrastive}, suggest that the congruent view with and without GD are expected to provide a global and local viewpoint when utilizing the underlying topological information. 
Therefore, in our model, we adopt a similar approach to assist G-Zoom 
in encoding rich local and global information of a graph simultaneously. The detailed explanation of 
GS, ED, GD, and NFD are shown below.

\subsubsection{Graph Sampling (GS)}
\label{section:IV-A(1)}
To ensure an 
augmented view includes all the neighbors, \changethird{i.e., all top-k strongly related neighbors,} of a batch of target nodes 
for generating neighboring-level embeddings, we have employed a special graph sampling approach. 
Given a batch with $B$ nodes, 
$k$ closely related neighbors to be sampled for each target node $v_t$, and the sampled subgraph size $P$, to generate the sampled graph $\hat{\mathcal{G}}$, we first need to include all $B$ target nodes and all of their neighbors, whose size is at most $B \times k$ in the sampled subgraph. In a batch, target nodes may share the same neighbors and thus, the number of $B$ target nodes plus their top-$k$ neighbors is likely to be smaller than $B \times k$. Then, we sample nodes from $\mathcal{G}$ excluding $B$ target nodes and their top-$k$ neighbors to enlarge the sampled graph to size $P$. In G-Zoom, to ensure all target nodes and their neighbors can be included in $\hat{\mathcal{G}}$, the sampled graph size $P$ has to be larger than the number of a batch of target nodes and their top-$k$ neighbors. 
Also, $P$ has to be smaller than the 
size of $\mathcal{G}$ to ensure that the sampled subgraph is included in the input graph. 
Then, we randomly sample nodes other than 
target nodes and their neighbors in the input graph $\mathcal{G}$ to complete 
$\hat{\mathcal{G}}$. 
An additional advantage of GS 
is enabling 
the batch processing to extend the scalability of our model, especially allowing the learning on large-scale graphs. 

\subsubsection{Edge Dropout (ED)}
Edge dropout is a type of augmentation placed on the graph topological space. 
With a predefined probability $p$, ED first randomly removes a 
$p$ percent of edges in a subgraph $\hat{\mathcal{G}}$, and then adds the same proportion of edges back to it. 
Please \changethird{note} that both adding and dropping edges process follow i.i.d. uniform distribution.

\subsubsection{Graph Diffusion (GD)}
We adopt graph diffusion for creating the incongruent view $\hat{\mathcal{G}_2}$ to inject 
global information to facilitate our different contrastive schemes. 
GD is proposed to solve two main issues in previous GNNs, which are the limited receptive field for message passing and 
noisy edges in real-world graphs \cite{klicpera2019diffusion}. Specifically, GD creates a new view $\tilde{\mathcal{G}}$ through the spatial message passing, which can distribute node information to a broader neighborhood instead of passing only to the first-hop neighbors. As such, GD enables the model to aggregate information from multi-hop 
neighbors and encode richer 
global information into 
nodes. This process can be formulated as follows:
\begin{equation}\label{eq: diff}
    \textbf{S} = \sum^{\infty}_{a = 0}\theta_{a}\textbf{T}^{a},
\end{equation}
where $\theta_a$ is a weighting term to determine the proportion of local and global information to be encoded. $\textbf{T}\in \mathbb{R}^{n \times n}$ denotes the generalized transition matrix. To ensure the convergence, two conditions including $\theta_a \in [0,1]$, $\sum^{+\infty}_{a = 0}\theta_{a} = 1$ and the eigenvalues of $\textbf{T}$ are restricted by $\lambda_i \in [0, 1]$ should be satisfied. 

\change{In our study, we adopted the Personalized PageRank (PPR) based graph diffusion $\textbf{S}^{PPR}$.} Given an adjacency matrix $\textbf{A} \in \mathbb{R}^{n \times n}$ and its degree matrix $\textbf{D} \in \mathbb{R}^{n \times n}$, the transition matrix $\textbf{T}$ and the weighting coefficient $\theta_a$ can be formulated as \change{$\textbf{T} = \textbf{D}^{-1/2}\textbf{A}\textbf{D}^{-1/2}$ and $\theta_a = \alpha(1 - \alpha)^a$}, respectively. Here $\alpha \in [0,1]$ denotes the teleport probability, which determines the tendency of returning to the starting node or teleporting to its neighbors in a random walk. \change{$\textbf{S}^{PPR}$ can be formulated as follows \cite{hassani2020contrastive}:}
\begin{equation}
    \textbf{S}^{PPR} = \alpha(\textbf{I} - (1 - \alpha)\textbf{D}^{-1/2}\textbf{A}\textbf{D}^{-1/2})^{-1}.
\label{eq: ppr_1}
\end{equation}
\change{Graph diffusion effectively generates an incongruent view with a more informative neighborhood. However, it is problematic when tackling large-scale graphs due to its memory-intensive operation on computing matrix inversion.} To overcome this problem and to make G-Zoom scalable to large-scale graphs, 
we have proposed a 
row-wise graph diffusion method, \change{ which is shown in Equation (\ref{eq: power-iteration-diff}). Specifically, we adopt power iteration (i.e., iteratively doing matrix multiplication) to approximate matrix inversion for diffusion matrix generation.}

\subsubsection{Node Feature Dropout (NFD)}
To augment 
\changesecond{the attributive information of a graph}, NFD randomly masking a fraction 
of feature dimensions in 
\changesecond{the feature matrix of $\mathcal{\hat{G}}$, i.e.,} $\hat{\textbf{X}}$, by defining an augmentation ratio $p$. In other words, after applying NFD, \changesecond{$p$ percent of the columns of $\hat{\textbf{X}}$} would be assigned zeroes.

After getting $\hat{\mathcal{G}_1}$ and $\hat{\mathcal{G}_2}$ via different graph augmentations, 
we feed them into a shared GNN encoder $g_{\theta}$. In our study, for simplicity, 
we adopt a 1-layer GCN \cite{kipf2016semi} as the backbone encoder: 
\begin{equation}\label{eq: gnn}
    \mathbf{H}_{\mathcal{G}}^{(l)} = \sigma(\hat{\textbf{D}}^{-\frac{1}{2}}\hat{\textbf{A}}\hat{\textbf{D}}^{-\frac{1}{2}}\textbf{H}^{(l-1)}_{\mathcal{G}}\textbf{W}^{(l-1)}) ,
\end{equation}
where $\hat{\textbf{A}} = \textbf{A} + \textbf{I}$, and $\hat{\textbf{D}}$ is the diagonal degree matrix of $\hat{\textbf{A}}$. In the above formula, $l$ denotes the $l$-th layer, and $\textbf{W}$ is a learnable weight matrix. We resort to this encoder to map node underlying topological and attributive information into the latent space, where we have node representations 
$\textbf{H}_1$ and $\textbf{H}_2$ for $\hat{\mathcal{G}_1}$ and $\hat{\mathcal{G}_2}$.

\subsection{Graph Contrastive Learning with Adjusted Zooming}
\label{section:IV-B}
In G-Zoom, we introduce 
a graph contrastive scheme based on the adjusted zooming, 
which injects contrastiveness among 
multiple scales: 
micro, meso, and macro, each of which represents 
the node-, neighboring-, and subgraph-level 
viewpoints. 

\subsubsection{Micro (node-level) contrastiveness}

\begin{figure}
\centering
\includegraphics[width=0.35\textwidth]{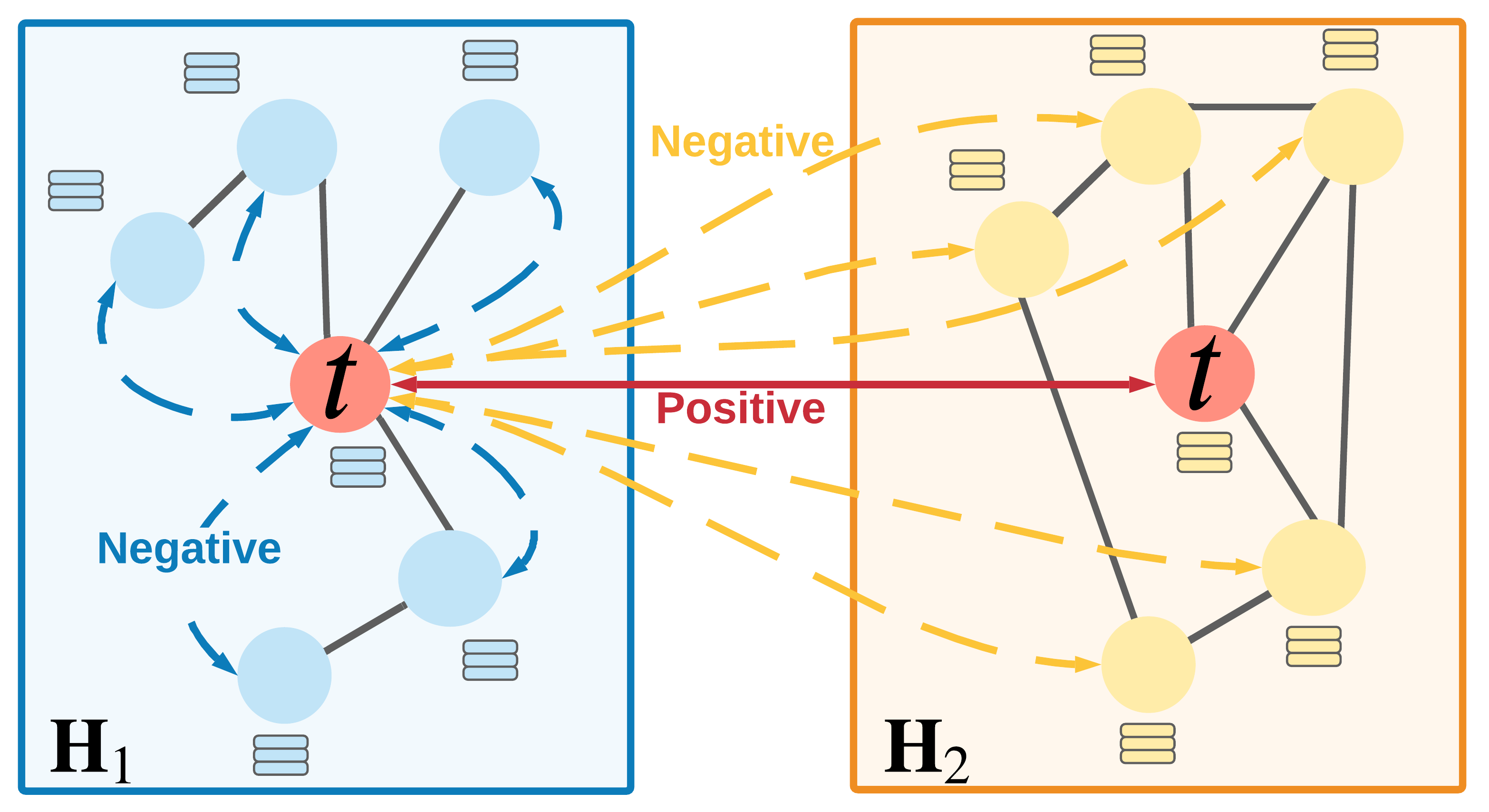} 
\caption{\small Contrasting from micro perspective compares the node representation of the target node $v_t$ from the representations of two graph views $\textbf{H}_1$ and $\textbf{H}_2$.}
\label{fig:3}
\end{figure}

Viewing a graph from the micro perspective helps G-Zoom focus on \changesecond{specific self-information of a node.} In this level of contrastiveness, we compare 
target node embeddings in two graph views, where each view contains the graph topological information on different scales.

After mapping the node features of $\hat{\mathcal{G}_1}$ and $\hat{\mathcal{G}_2}$ to the latent space \change{via the GNN encoder $g_{\theta}$}, 
we have $\textbf{H}_1$ and $\textbf{H}_2$ as the node representation matrices of two graph views. \change{This GNN encoder can be replaced with different GNN methods (e.g., GCN~\cite{kipf2016semi} and GAT~\cite{velivckovic2017graph}). For simplicity, we only use a 1-layer GCN as the backbone graph encoder in our model.}
As shown in Figure \ref{fig:3}, given a target node $v_t$, we first maximize the cosine similarity of the positive pair $(\textbf{h}_1^t, \textbf{h}_2^t)$, i.e., the red solid line, where $\textbf{h}_1^t \in \textbf{H}_1$ and $\textbf{h}_2^t \in \textbf{H}_2$ are the 
representations of $v_t$ in two views.
With this maximization, we pull the representations of a node closer in two views to learn the invariant patterns.
To regularize this discrimination and avoid model collapse, 
we have defined two negative sampling strategies to consist of our micro contrastiveness, which further exploits the rich contrastive relations within and between $\textbf{H}_1$ and $\textbf{H}_2$. In Figure \ref{fig:3}, the blue dash line shows the first 
type of negative sampling, 
which considers all nodes except the target node within a view as negative samples. Similarly, the yellow dash line indicates 
the second type of negative sampling, which 
regards all nodes excluding the target node in a different view as negative counterparts. The cosine similarity of all these negative pairs needs to be minimized. Thus, the aforementioned process can be formulated as follows:

\begin{equation}\label{eq:micro_view_1}
\mathcal{L}_{(\textbf{h}^t_1, \textbf{h}^t_2)} = \log\frac{e^{s(\textbf{h}^t_1, \textbf{h}^t_2)}}{e^{s(\textbf{h}^t_1, \textbf{h}^t_2)} + \sum^2_{v=1}\sum^B_{i=1;i \neq t}e^{s(\textbf{h}^t_1, \textbf{h}^i_{v})}},
\end{equation}
where $s(\cdot)$ is the cosine similarity, 
$B$ denotes the number of target nodes in a batch
, and $v \in [1,2]$ denotes one of the augmented view. \changesecond{If $v=1$, $\textbf{h}^i_{v}$ represents the representation of node $i$ in $\textbf{H}_1$, whereas when $v=2$, $\textbf{h}^i_{v}$ is the node representation of node $i$ in $\textbf{H}_2$.} \changesecond{Taking the argument in an opposite position, i.e., $(\textbf{h}_2^t,\textbf{h}_1^t)$, the loss $\mathcal{L}_{(\textbf{h}_2^t,\textbf{h}_1^t)}$ is combined with $\mathcal{L}_{(\textbf{h}_1^t,\textbf{h}_2^t)}$ to reinforce the self-supervised signals. The micro-level objective function of our model can be defined as follows:} 
\begin{equation}\label{eq:micro}
\mathcal{L}_{micro} = -\frac{1}{2B}\sum^B_{\change{t = 1}}\big(\mathcal{L}_{(\textbf{h}^t_1, \textbf{h}^t_2)} + \mathcal{L}_{(\textbf{h}_2^t,\textbf{h}_1^t)}\big).
\end{equation}

\subsubsection{Meso (neighboring-level) contrastiveness}
\label{meso(neighboring-level) contrastiveness}
Unlike exploring a graph from a micro or macro perspective, which 
only contrasts from a fixed viewpoint (i.e., zooming scale) 
, examining a graph from the meso perspective is 
more flexible
, which means we can adaptively select a zooming scale \changesecond{that} 
lies between the finest and the broadest viewpoint. Before we conduct contrasting 
from the meso perspective, we select a predefined number $k$ of the most influential neighbors for each target node $v_t \in \mathcal{G}$ by calculating the importance between them. 
Specifically, this process can be conducted offline and in parallel without compromising the efficiency of model training and inference. 
To achieve this, there are two scoring approaches to measure the importance among nodes in a graph: 
the random walk (RW)-based and Personalized Page Rank (PPR)-based approach. The first method determines \changesecond{the importance of a node $v_i$} to the target node $v_t$ by creating a randomly generated sequence starting from node $v_t$ and \changesecond{counting the occurrence frequency of the node $v_i$ in the sequence.} 
The higher the occurrence frequency of $v_i$, the more important it is to $v_t$.
On the other hand, the latter approach utilizes the PPR algorithm, whose formulation is shown in Equation (\ref{eq: ppr_1}). In G-Zoom, we adopt a node-wise PPR approach for the importance calculation between nodes. Instead of conducting the PPR operation on a whole graph at once, our approach computes the node importance row by row. Compared with the RW
-based method, our approach can be naturally paralleled and has an advantage in computation efficiency. 

\begin{figure}[t]
\centering
\includegraphics[width=0.4\textwidth]{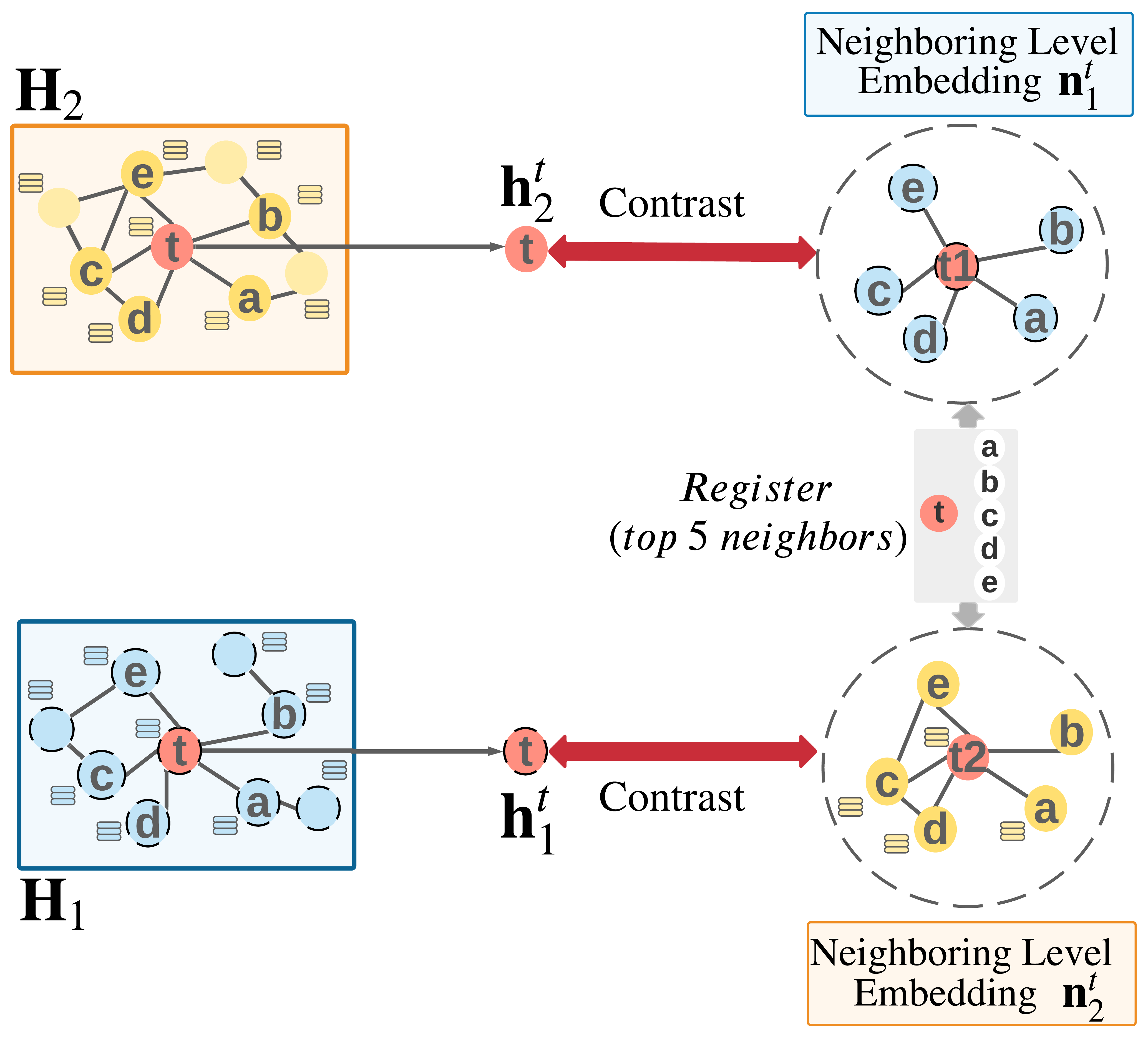}
\caption{\small Contrasting from meso perspective compares the node representation of the target node $v_t$ from one view with the neighboring-level embeddings in the other view. The neighboring-level representation is generated based on a neighborhood register. Here the red line denotes the contrastiveness.}
\label{fig:4}
\end{figure}

As shown in Figure \ref{fig:4}, to generate the neighboring-level representations 
for a batch of sampled target nodes, we design 
a neighborhood register, which stores a list of top-$k$ important neighbors for each target node, 
where $0 \leq k \leq P, k \in \mathbb{Z}$ is a tunable hyperparameter \changesecond{for selecting the number of neighbors for a target node}, which decides the discrimination scope of our meso-based contrastiveness. \change{The definition of the neighborhood register is \changethird{as} follows:}
\change{
\begin{definition}[Neighborhood Register]
\label{def:register}
 We define the neighborhood register \changesecond{$\textbf{R} \in \mathbb{R}^{n \times k}$} as a collection of \changesecond{$\textbf{r}_i \in \mathbb{R}^{\changethird{k}}$}, which stores the index of top $k$ important neighbors of node $i$ retrieved from the importance matrix $\textbf{I}$.
\end{definition}}
With this register, we can calculate the neighboring-level representations for target nodes in a batch by \change{averagely} aggregating the node representations in the register via the pooling layer \change{(i.e., mean pooling)}.

In Figure \ref{fig:4}, 
the neighborhood register includes the top-5 strongly correlated neighbors for the target node $v_t$, namely $v_a$, $v_b$, $v_c$, $v_d$, and $v_e$.  
With this register, we can calculate the neighboring-level representations for target nodes in a batch by aggregating the node representations in the register via the pooling layer. Specifically, for a target node $v_t$, we use $\textbf{n}^t_1$ and $\textbf{n}^t_2$ to denote its neighboring-level representations in $\textbf{H}_1$ and $\textbf{H}_2$. 
This aggregation process is presented as the yellow and blue boxes in the figure. Then, to establish the neighboring-level contrastiveness, we use the idea of Deep InfoMax \cite{hjelm2018learning} to maximize the mutual information between two positive pairs $(\textbf{h}_1^t, \textbf{n}_2^t)$ and $(\textbf{h}_2^t, \textbf{n}_1^t)$, while pushing 
away two negative pairs $(\tilde{\textbf{h}^t}, \textbf{n}_2^t)$ and $(\tilde{\textbf{h}^t}, \textbf{n}_1^t)$ by 
minimizing their agreement,  \change{where $\tilde{\textbf{h}}^t$ is the corrupted representation of $v_t$ extracted from the learned representations $\tilde{\textbf{H}}$ of a randomly shuffled feature matrix $\tilde{\textbf{X}}$. The contrastive objective for $(\textbf{h}_1^t, \textbf{n}_2^t)$ can be formulated as}:
\begin{equation}
    \mathcal{L}_{(\textbf{h}_1^t, \textbf{n}_2^t)} = \sum_{t=1}^B \log\mathcal{D}(\textbf{h}_1^t, \textbf{n}_2^t) + \log(1 - \mathcal{D}(\tilde{\textbf{h}^t}, \textbf{n}_2^t)),
\label{eq: loss_1}
\end{equation}
\change{\noindent where $\mathcal{L}_{(\textbf{h}_1^t, \textbf{n}_2^t)}$ denotes the neighboring-level contrastiveness loss for $(\textbf{h}_1^t, \textbf{n}_2^t)$.} Also, $B$ is the number of nodes in a batch, and $\mathcal{D}$ is a discriminator, which computes the agreement between the elements in an instance pair to evaluate its mutual information.   \changesecond{The formulation of $\mathcal{D}$ is shown in Equation (\ref{eq:discriminator}). Similarly, we can obtain the contrastive loss $\mathcal{L}_{(\textbf{h}_2^t, \textbf{n}_1^t)}$.} 
The overall neighboring-level contrastiveness can be defined by marrying $\mathcal{L}_{(\textbf{h}_1^t, \textbf{n}_2^t)}$ and $\mathcal{L}_{(\textbf{h}_2^t, \textbf{n}_1^t)}$:
\begin{equation}\label{eq:meso}
    \mathcal{L}_{meso} = -\frac{1}{2B}(\mathcal{L}_{(\textbf{h}_1^t, \textbf{n}_2^t)} + \mathcal{L}_{(\textbf{h}_2^t, \textbf{n}_1^t)}).
\end{equation}

\subsubsection{Macro (subgraph-level) contrastiveness}
From the macro perspective, our model can have a bird eye view to inspect 
all nodes within a sampled graph. At this level, we define a contrastive path 
by maximizing the agreement between target node embeddings 
and subgraph-level representations, which can be obtained by aggregating all node embeddings in the sampled subgraph with size $P$, which consists of a batch of target nodes, their top-$k$ neighbors, and a small portion of randomly sampled nodes. 
This process is expected to extract 
rich global information from the graph.
\begin{figure}
\centering
\includegraphics[width=0.4\textwidth]{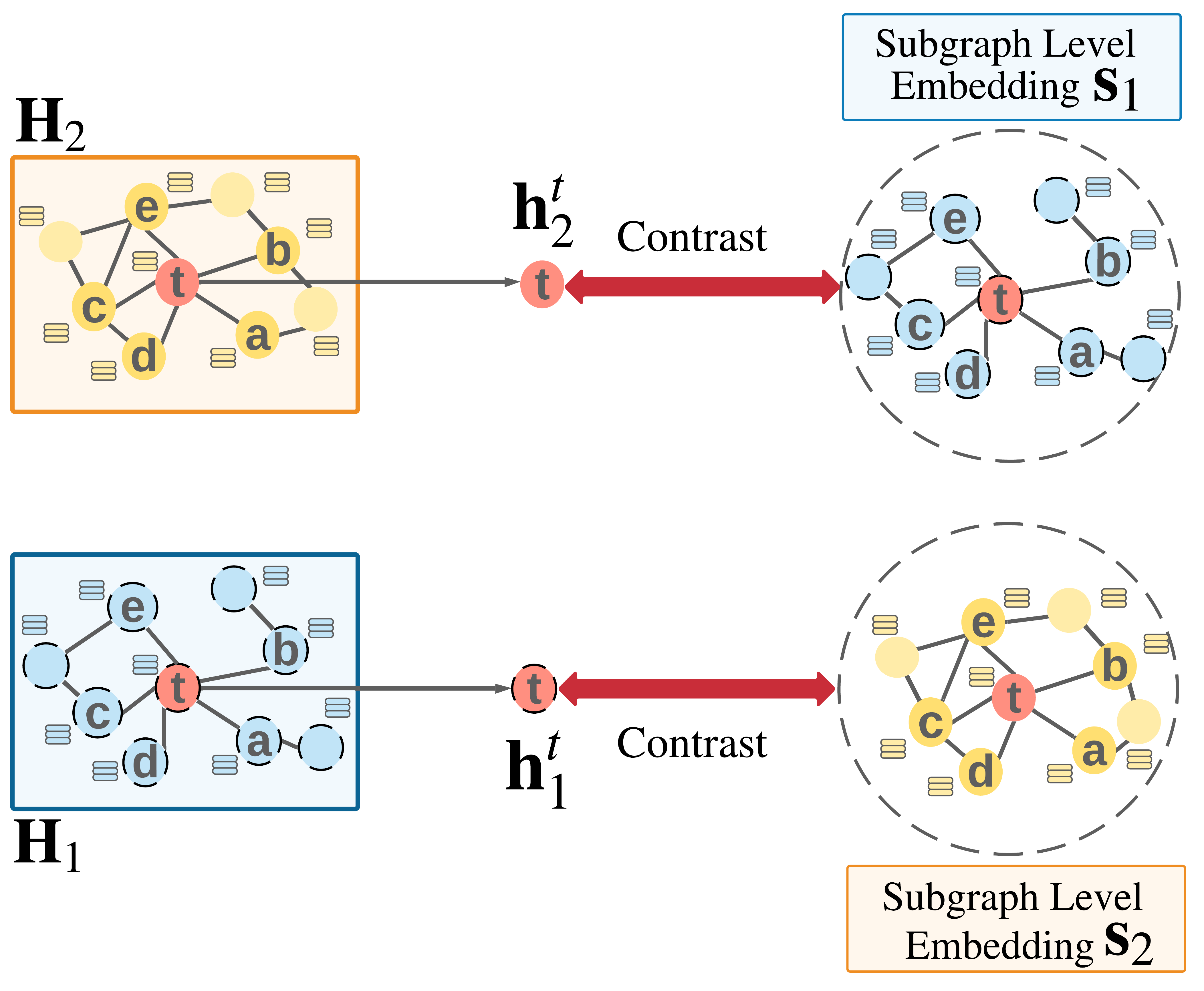}
\caption{\small Contrasting from macro perspective compares the node representation of the target node $v_t$ from one view with the subgraph-level embeddings in the other view. Here the red line denotes the contrastiveness.}
\label{fig:5}
\end{figure}
As shown in Figure \ref{fig:5}, to obtain the subgraph-level representations (i.e., $\textbf{s}_1$ for $\textbf{H}_1$ and $\textbf{s}_2$ for $\textbf{H}_2$), we fed $\textbf{H}_1$ and $\textbf{H}_2$ 
into the 
pooling layer\changethird{.} 
Then, we contrast the target node embeddings with subgraph-level representations between different views, i.e., 
$\textbf{h}_1^t$ versus $\textbf{s}_2$ and $\textbf{h}_2^t$ versus $\textbf{s}_1$. 
The contrastive 
loss for the instance 
pair ($\textbf{h}_1^t, \textbf{s}_2$) can be formulated as:

\begin{equation}
    \mathcal{L}_{(\textbf{h}_1^t, \textbf{s}_2)} = \sum_{t=1}^B \log\mathcal{D}(\textbf{h}_1^t, \textbf{s}_2) + \log(1 - \mathcal{D}(\tilde{\textbf{h}^t}, \textbf{s}_2)),
\end{equation}
and \changesecond{we can calculate the loss $\mathcal{L}_{(\textbf{h}_2^t, \textbf{s}_1)}$ in a similar way.} 
By merging these two losses, we can get the objective function of our subgraph-level contrastiveness: 
\begin{equation}\label{eq:macro}
    \mathcal{L}_{Macro} = -\frac{1}{2B}(\mathcal{L}_{(\textbf{h}_1^t, \textbf{s}_2)} + \mathcal{L}_{(\textbf{h}_2^t, \textbf{s}_1)}).
\end{equation}

\subsection{Model Optimization and Algorithm}
\label{section:IV-C}
\begin{equation}
    \mathcal{L} = \alpha*\mathcal{L}_{micro} + \beta*\mathcal{L}_{meso} + \gamma*\mathcal{L}_{macro}.
    \label{eq: loss_overall}
\end{equation}

\IncMargin{1em}
\begin{algorithm}
\SetKwData{Left}{left}\SetKwData{This}{this}\SetKwData{Up}{up}
\SetKwFunction{Union}{Union}\SetKwFunction{FindCompress}{FindCompress}
\SetKwInOut{Input}{Input}\SetKwInOut{Output}{Output}
\Input{Input Graph: $\mathcal{G}$; Diffused Graph: $\Tilde{\mathcal{G}}$; Batch Size: $B$; Number of top neighbors for target node $v_t$: $k$; Neighborhood Register: $\textbf{R}$; Sample Size: $P$; Maximum training epochs\changethird{:} $E$;}
\Output{Final representations $\Vec{\textbf{H}}$}
\BlankLine
\emph{//Model Training}\;
\For{$e = 1$ to $E$}{
\emph{//Batch sampling}\;
Sample $B$ target nodes \changethird{$V$} to create a batch\;
Create $\hat{\mathcal{G}}_1$ and $\hat{\mathcal{G}}_2$ from $\mathcal{G}$ and $\mathcal{\tilde{G}}$, respectively, as described in section \ref{section:IV-A(1)}\;
\emph{//Obtain node-, neighboring-, and subgraph-level representations}\;
\For{\changethird{$v_t$ in $V$}}{
Obtain $\textbf{h}^t_1$ and $\textbf{h}^t_2$ via \change{Equation (\ref{eq: gnn})}\;
Obtain $\textbf{n}^t_1$ and $\textbf{n}^t_2$ by aggregating neighbors stored in $\textbf{R}$\;
Obtain $\textbf{s}_1$ and $\textbf{s}_2$ by aggregating all nodes in $\hat{\mathcal{G}}_1$ and $\hat{\mathcal{G}}_2$\;
}
\emph{//Compute Loss}\;
Calculate $\mathcal{L}_{micro}$ via \change{Equation (\ref{eq:micro})}\;
Calculate $\mathcal{L}_{meso}$ via \change{Equation (\ref{eq:meso})}\;
Calculate $\mathcal{L}_{macro}$ via \change{Equation (\ref{eq:macro})}\;
Calculate $\mathcal{L}$ via \change{Equation (\ref{eq: loss_overall})}\;
\emph{//Update parameters}\;
Update trainable matrix $\textbf{W}$ in encoder $g_{\theta}$\ and $\textbf{W}_D$ in discriminator shown in \change{Equation (\ref{eq:discriminator})};
}
\emph{//Inference}\;
Obtain final embeddings $\Vec{\textbf{H}} =g_{\theta}(\mathcal{G})+g_{\theta}(\tilde{\mathcal{G}})$
\caption{The Overall Procedure of G-Zoom}\label{algo_disjdecomp}
\end{algorithm}\DecMargin{1em}
To train our model and learn effective node embeddings for downstream tasks, we combine the contrastiveness loss from the aforementioned three perspectives to define the overall objective function of G-Zoom, 
where $\alpha$, $\beta$ and $\gamma$ are three weighting terms for controlling the contribution of each contrastiveness loss in model training. During the training process, we aim to minimize the overall loss $\mathcal{L}$. For model inference, 
to obtain the final representation of a target node $v_t$, we combine 
$\textbf{h}_1^t$ and $\textbf{h}_2^t$, i.e.,  $\textbf{h}_1^t$ + $\textbf{h}_2^t \in \mathbb{R}^{\changethird{d'}}$ to get the final node representation $\vec{\textbf{h}^t}$.

\noindent\textbf{Algorithm} The overall procedure of G-Zoom 
is depicted in Algorithm 1. From Algorithm 1, we can see the overall procedure consists of 
two stages: model training and inference. 
During model training, within every epoch, we create two augmented graph views 
$\hat{\mathcal{G}}_1$ and $\hat{\mathcal{G}}_2$ with the sample size $P$ as the inputs of our 
GNN encoder $g_{\theta}$ in 
three steps. Firstly, we sample 
$B$ target nodes as the base of graph views. 
Then, we put all these target node top-$k$ neighbors in the sampled views. 
Finally, we enlarge these two views 
to size $P$ by adding randomly sampled nodes from the input graph $\mathcal{G}$ excluding the selected $B$ target nodes and their top-$k$ neighbors.

After this, we feed $\hat{\mathcal{G}}_1$ and $\hat{\mathcal{G}}_2$ to $g_{\theta}$, which outputs 
$\textbf{H}_1$ and $\textbf{H}_2$. Then, by leveraging the 
neighborhood register and pooling layer, we construct three different contrastive paths in the latent space from various perspectives (e.g., micro, meso, and macro). 
Finally, 
we have our 
overall loss $\mathcal{L}$ by 
combining three levels of contrastiveness. After this, 
a backward propagation with the 
gradient descent algorithm fuel the parameter updating 
process for $g_{\theta}$.

At the inference stage, 
summing up the result representations of input graph $\mathcal{G}$ and the diffused graph $\tilde{\mathcal{G}}$ generates the final embedding matrix $\vec{\textbf{H}}$.

\section{Complexity Analysis and Further Extension}
We provide an analysis of our  algorithm G-Zoom and discuss how to extend G-Zoom to handle large-scale graphs. 
\subsection{Time complexity of G-Zoom}
\label{section:IV-D}
In this section, we analyze the time complexity of G-Zoom by considering \change{two main components: graph augmentation and the contrastive learning process}. Specifically, we train our model at most $E$ times, and within every epoch, we conduct subsampling on the input graph once, whose time complexity is $O(P)$, where $P$ is the sample size. Therefore the total complexity for this component is $O(EP)$. 

\change{For the contrastive learning process, its time complexity is mainly contributed by the GNN encoder, the infoNCE loss computation and the discriminator.} The time complexity of the GNN encoder is $O(\mathcal{E})$ \change{with a sparse aware matrix product, where $\mathcal{E}$ is the number of edges of a graph. In our method, with a 1-layer GCN, the time complexity of the GNN component is $O(\mathcal{E}_1 + \mathcal{E}_2)$, where $\mathcal{E}_1$ is the number of edges for the original sampled graph, and $\mathcal{E}_2$ represents the diffused one. With diffusion, $\mathcal{E}_2$ is significantly larger than $\mathcal{E}_1$.}  
\change{Then, we compute the infoNCE loss for micro-level contrastiveness, whose time complexity is $O(B^2d')$, as shown in Equation (5), where $B$ is the batch size of target nodes. 
As shown in Equation (13), we adopt a bilinear discriminator for calculating the other two losses (i.e., meso-, and macro-level). In our methods, in each training epoch, we need to process the two-level losses using Equation (13). As shown in Equation (7) and Equation (9), each of these two losses requires $O(2Bd')$ for computation. Therefore, calculating these two-level losses requires $O(4Bd')$. 
To sum up, in each training epoch, the time complexity of processing all three contrastiveness is $O(4Bd' + 2B^2d')$. 
Also, \changesecond{as shown in Equation (\ref{eq:micro}), (\ref{eq:meso}) and (\ref{eq:macro}), we combine two losses to reinforce the self-supervised signals}, the time complexity of the computation should be doubled to $O(8Bd' + 4B^2d')$.}
\change{By combining the two main components, the overall complexity is $O(E(P + \mathcal{E}_1 + \mathcal{E}_2 + 8Bd' + 4B^2d'))$. As we adopted Big O notation, we ignore the constant, and the final time complexity becomes $O(E(P + \mathcal{E}_1 + \mathcal{E}_2 + Bd' + B^2d')$).}

\subsection{Extending G-Zoom to large-scale graphs}
\label{section:IV-E}
To fulfill the applications in the real world and increase the scalability of G-Zoom, 
we extend our proposed algorithm to large-scale graphs. Most self-supervised GNN studies cannot handle large input graphs mainly due to two reasons. Firstly,  
they feed the entire graph as the input 
for model training, e.g., DGI\cite{velickovic2019deep}, GRACE\cite{zhu2020deep} and GMI\cite{peng2020graph}. In this case, the capacity of a model is constrained 
by the input graph size. 
Secondly, 
some of them adopt a non-expandable graph diffusion approach as their 
structural graph augmentation, e.g., MVGRL\cite{hassani2020contrastive}. 

To overcome the aforementioned two issues, 
\change{we first use a tailored sub-sampling sampler, which has been introduced in subsection IV-A (1), to create subgraphs. This sampler not only alleviates the restriction of the input graph size in model training but can also sufficiently include the important features of the initial graph. To prove the effectiveness of this sampler, we present the theoretical analysis in Appendix A.}

\change{To tackle the second issue, we apply a scalable graph diffusion method to create a 
diffused graph $\tilde{\mathcal{G}}$ and then use the aforementioned sub-sampling strategy to generate the augmented view $\hat{\mathcal{G}}_2$. Without additional graph diffusion operations, $\hat{\mathcal{G}}_1$ is obtained 
by graph sampling.} Specifically, we adopt a row-wise power-iteration 
based diffusion approach with the PPR kernel 
which instead of getting the final diffused graph at once by applying \change{Equation
(\ref{eq: ppr_1})}, we calculate the importance score node-by-node 
and then concatenate all output importance vectors together to form the final diffused graph. 
There are two advantages of applying this row-wise diffusion approach: 
\begin{itemize}
  \item Parallel the computation of diffusion matrix, which significantly speeds up the diffusion process on graphs.
  \item Alleviate the resource burdens of matrix inversion when calculating the graph diffusion matrix, which only supports the operations on dense matrices.
\end{itemize}
With power iteration, we can approximate the final diffusion matrix 
by iteratively conducting the 
matrix multiplication rather than computing matrix inversion. This process can be formulated as: 
\begin{equation}
\label{eq: power-iteration-diff}
    \pi_{ppr}(v_t) = (1 - \alpha)i + \alpha \tilde{\textbf{A}}\pi_{ppr}(v_t),
\end{equation}
where $v_t$ is a node, $i$ is a one-hot identity vector, in which only the $i$-th row entry is one, and all remaining entries are 0, $\pi_{ppr}(v_t)$ is the importance vector that stores all nodes' importance to $v_t$, $\alpha$ is a teleport probability determining the tendency of teleporting back to node $t$, and $\tilde{\textbf{A}}$ is the normalized adjacency matrix $\textbf{A}\textbf{D}^{-1}$, where $\textbf{D}$ is the diagonal matrix. \changesecond{To show the impact of employing the subsampling and the proposed scalable graph diffusion method, we compare the performance of G-Zoom with two variants, which remove the subsampling or use the scalable graph diffusion in Section \ref{section:V-D}}.

\section{Experiment}
\label{section:VI}
In this section, we have conducted experiments to present the effectiveness of G-Zoom. Firstly, we introduce five benchmark datasets 
and 
experiment setup in Subsection \ref{section:V-A} and \ref{section:V-B}. After this, 
the overall comparisons, ablation study, parameter sensitivity analysis, and visualization are illustrated from Subsection \ref{section:V-C} to \ref{section:V-G}.

\subsection{Datasets}
\label{section:V-A}
We have selected five benchmark datasets varying with sizes, including Cora, Citeseer, Coauthor CS, Coauthor Physics, and ogbn-arixv for experiments. Within each dataset, nodes represent scientific publications and are connected via citations. As for attributive information
, while the first four datasets carry the bag-of-word representation, \changesecond{the feature vector of ogbn-arxiv} is obtained by averaging the word embeddings of paper title and abstract. The statistics of this dataset are summarized in Table \ref{tab: dataset statistics}. 

\begin{table}[t]
	\small
	\caption{The statistics of five benchmark datasets.}
	\centering
	{
    	\begin{tabular}{llcccc}
    		\toprule
    		Dataset & Nodes & Edges & Features & Classes \\
    		\midrule
    			\textbf{Cora} & 2,708 & 5,429 & 1,433 & 7\\
    			\textbf{Citeseer} & 3,327 & 4,732 & 3,703 & 6\\
    			\textbf{Coauthor CS} & 18,333 & 81,894 & 6,805 & 15\\
    			\textbf{Coauthor Physics} &	34,493 & 991,848 & 8,415& 5\\
    			\textbf{ogbn-arxiv}	& 169,343 & 1,166,243 & 128 & 40\\
    		\bottomrule
    	\end{tabular}
    }
\label{tab: dataset statistics}
\end{table}

As shown in the table above, the first two datasets only have thousands of nodes and edges, whereas Coauthor CS and Coauthor Physics \changethird{have} over ten thousand vertices and edges. Retrieving from Open Graph Benchmark (OGB)\cite{hu2020open}, ogbn-arxiv is a large-scale graph dataset with over a hundred thousand nodes and a million edges, which is typically used for evaluating the robustness and scalability of the graph models.

\subsection{Experiment Setup}
\label{section:V-B}
\begin{table*}[htbp]
	\small
	\caption{Overall performance comparison of nine baselines and G-Zoom on five benchmark datasets. \textbf{X, A} and \textbf{Y} \changethird{represent} whether the method utilizes feature matrix, adjacency matrix, and label information, respectively. OOM denotes 
	OUT-OF-MEMORY. The highest accuracy value is in bold for each dataset.}
	\centering
	{
    	\begin{tabular*}{1.0\textwidth}{@{\extracolsep{\fill}}llccccc}
    		\toprule
    		Information Used & Method & Cora & CiteSeer & Coauthor Physics & Coauthor CS & ogbn-arxiv \\
    		\midrule
    			\textbf{X, A, Y} & GCN & 81.5 & 70.3   & 95.7 $\pm{0.2}$  & 93.0 $\pm{0.3}$ & 68.15 $\pm{0.1}$ \\
    			\textbf{X, A, Y} & GAT & 83.0 $\pm{0.7}$ & 72.5 $\pm{0.7}$   & 95.5 $\pm{0.2}$  & 92.3 $\pm{0.2}$ & 68.85 $\pm{0.1}$ \\
    			\textbf{X, A, Y} & SGC & 81.0 $\pm{0.0}$ & 71.9 $\pm{0.1}$ & 95.8 $\pm{0.1}$  & 92.7 $\pm{0.1}$ & 68.91 $\pm{0.1}$ \\
    			\textbf{X, A, Y} & CG3 & 83.4 $\pm{0.7}$ & 73.6 $\pm{0.8}$ & OOM  & 92.3 $\pm{0.2}$ & OOM \\
    		\midrule
    			\textbf{X, A}	& DGI & 81.7 $\pm{0.6}$ & 71.5 $\pm{0.7}$  & 94.5 $\pm{0.5}$  & 92.2 $\pm{0.6}$ & OOM \\
    			\textbf{X, A}	& GMI & 82.7 $\pm{0.2}$ & 73.0 $\pm{0.3}$  & OOM & OOM  & OOM \\
    			\textbf{X, A}	& MVGRL & 82.9 $\pm{0.7}$ & 72.6 $\pm{0.7}$  & 95.3 $\pm{0.1}$  & 92.1 $\pm{0.1}$ & OOM \\
    			\textbf{X, A}	& GRACE & 80.0 $\pm{0.4}$ & 71.7 $\pm{0.6}$  & \change{95.2 $\pm{0.2}$} & 92.8 $\pm{0.1}$ & OOM\\
    			\textbf{X, A}	& SubG-Con & 83.5 $\pm{0.5}$ & 73.2 $\pm{0.2}$  & 96.2 $\pm{0.4}$  & 94.0 $\pm{0.6}$ & 55.7 $\pm{0.2}$ \\
    		\midrule
    			\textbf{X, A}	& G-Zoom & \textbf{84.7} $\bm{\pm{0.4}}$ & \textbf{74.2} $\bm{\pm{0.3}}$ & \textbf{96.6} $\bm{\pm{0.3}}$  & \textbf{94.9} $\bm{\pm{0.3}}$ & \textbf{70.1} $\bm{\pm{0.1}}$ \\
    		\bottomrule
    	\end{tabular*}
    \label{tab: Comparison Result}
	}
\label{tab: classification results}
\end{table*}

\begin{table*}[htbp]
	\small
	\caption{Performance comparison of G-Zoom and its six variants 
	on five benchmark datasets.}
	\centering
    	\begin{tabular*}{1.0\textwidth}{@{\extracolsep{\fill}}llcccc}
    		\toprule
    		Loss & Cora & Citeseer & Coauthor CS&Coauthor Physics & ogbn-arxiv\\
    		\midrule
    			$\textbf{G-Zoom}_{Macro}$ & 83.4 $\pm{0.4}$& 73.9 $\pm{0.4}$ & 92.9 $\pm{0.7}$ & 96.0 $\pm{0.3}$& 69.4 $\pm{0.1}$\\
    			$\textbf{G-Zoom}_{Meso}$ & 83.7 $\pm{0.7}$ & 73.9 $\pm{0.6}$ & 93.1 $\pm{0.6}$ & 95.6 $\pm{0.2}$ & 69.4 $\pm{0.1}$ \\
    			$\textbf{G-Zoom}_{Micro}$ & 77.3 $\pm{1.2}$ & 66.7 $\pm{1.1}$ & 89.2 $\pm{0.9}$ & 95.6 $\pm{0.3}$ & 69.9 $\pm{0.1}$\\
    		\midrule
    			$\textbf{G-Zoom}_{w/o Micro}$ & 83.7 $\pm{0.5}$ & 73.8 $\pm{0.4}$ & 92.8 $\pm{0.4}$& 95.4 $\pm{0.3}$&  69.5 $\pm{0.1}$\\
    			$\textbf{G-Zoom}_{w/o Macro}$ & 84.3 $\pm{0.7}$ & 73.3 $\pm{0.4}$ & 93.9 $\pm{0.6}$& 96.3 $\pm{0.3}$ &  70.1 $\pm{0.1}$\\
    			$\textbf{G-Zoom}_{w/o Meso}$ & 84.4 $\pm{0.5}$ & 73.2 $\pm{0.4}$ & 94.0 $\pm{0.5}$& 96.2 $\pm{0.2}$ & \textbf{70.2}  $\bm{\pm{0.1}}$\\
    		\midrule
    			$\textbf{G-Zoom}$ & \textbf{84.7} $\bm{\pm{0.4}}$ & \textbf{74.2} $\bm{\pm{0.3}}$  & \textbf{94.9} $\bm{\pm{0.3}}$  & \textbf{96.6} $\bm{\pm{0.3}}$ & 70.1 $\pm{0.1}$ \\
    		\bottomrule
    	\end{tabular*}
\label{tab: Ablation Study}
\end{table*}

In this section, we discuss specific module designs and hyper-parameter settings in G-Zoom. 
In our experiment, for simplicity
, we adopt a 1-layer 
GCN 
as the backbone graph encoder in our model, as shown in \change{Equation (\ref{eq: gnn})}. 
It is worth noting that 
our graph 
encoder can also be replaced by other GNN-based candidate models, such as GAT and SGC.



For the discriminator used in meso and macro contrastiveness, we consider a simple bilinear scoring function 
\cite{oord2018representation} as the discriminator $\mathcal{D}$ in our framework, which can be formulated as: 
\begin{equation}
    \mathcal{D}(\textbf{h}_i, \textbf{x}_j) = \sigma(\textbf{h}_i^T \textbf{W}_D \textbf{x}_j),
    \label{eq:discriminator}
\end{equation}
where $(\textbf{h}_i \in \mathbb{R}^{\changethird{d'}}, \textbf{x}_j \in \mathbb{R}^{\changethird{d'}})$ are a pair of representation to be scored, $\sigma$ is a non-linear activation function and $\textbf{W}_D \in \mathbb{R}^{d' \times d'}$ represents a trainable matrix. 

For hyper-parameter settings, all experiments are implemented by using PyTorch \cite{paszke2019pytorch}. In the training phase, we set the batch size $B$ to 200 and 400 for Cora and Citeseer, 
while $B$ for the three remaining datasets are set to $500$. For the number of strongly related neighbors $k$ of $B$ target nodes to be collected in the sampled graph, we define $k$ as 100 for all datasets except ogbn-arxiv where $k=10$. 
As discussed in Subsection \ref{section:IV-A}(1), the sample graph size $P$ has to be larger than the number of $B$ and their top-k neighbors. As such, $P$ is set to be larger than $B \times k$ to ensure $P$ \changethird{meets} this requirement. Specifically, we set 3000 for the first two datasets, while $P$ is 7000 for Coauthor CS, Coauthor Physics, and ogbn-arxiv. To improve the training efficiency, we employ an early stopping mechanism to detect the model convergence, 
and halt the training process based on a predefined patience value. 
The training epochs are fixed to at most 3000 for all datasets. The patience is defined as 50 for the first two datasets and 
200 for Coauthor datasets 
and ogbn-arxiv. Except for ogbn-arxiv, whose learning rate is 0.0001, all datasets have the learning rate tuned to 0.001. Also, for the weighting term $\alpha$, $\beta$, and $\gamma$ used in loss calculation as presented in \change{Equation (\ref{eq: loss_overall})}, they are all set to 1 except for Coauthor Physics, where $\alpha$ is set to 0.8. Therefore, in the experiment, the three-level contrastiveness contributes almost equally to our model training.

\subsection{Performance Comparison of G-Zoom and Baselines}
\label{section:V-C}
In this subsection, we compare the performance of G-Zoom 
in aforementioned five datasets with nine baselines, including four supervised 
and five self-supervised baselines. We adopt linear node classification as the evaluation protocol to validate the expressiveness of the learned node representations. 
\subsubsection{\textbf{Supervised GRL Baselines}}
\begin{itemize}
    \item GCN \cite{kipf2016semi} Graph Convolution Network (GCN) adapts the traditional convolution network used in image processing to graph-structured data. 
    \item GAT \cite{velivckovic2017graph} Graph Attention Network (GAT) adopts an attention mechanism to consider weights of neighbors in information aggregation instead of simply averaging their representations.
    \item SGC \cite{wu2019simplifying} Simple Graph Convolution (SGC) reduces the complexity of GCN by removing its nonlinearities and collapsing weight matrices.
    \item CG3 \cite{wan2020contrastive} CG3 utilizes both data similarities and graph structure to enrich the supervision signals in Graph-based Semi-Supervised Learning.
\end{itemize}

\subsubsection{\textbf{Self-supervised GRL Baselines}}
\label{section:V-D}
\begin{itemize}
    \item DGI \cite{velickovic2019deep} Deep Graph Infomax (DGI) adapts the idea of Mutual Information (MI) maximization from image processing \cite{hjelm2018learning} to graphs. Specifically, DGI implements a patch-summary maximization strategy (i.e., maximizing MI between node- and graph-level representation).
    \item GMI \cite{peng2020graph} Graph Mutual Information (GMI) focuses on maximizing MI between input graphs and high-level hidden representations based on node features and topological structure.
    \item MVGRL \cite{hassani2020contrastive} Multi-view Graph Representation Learning (MVGRL) utilizes graph augmentation techniques to generate multiple views and maximize the mutual information of the output node-level and readout subgraph-level representations between different views.
    \item GRACE \cite{zhu2020deep} Graph Contrastive Representation Learning (GRACE) also applies graph augmentation for creating two augmented views. Unlike the aforementioned self-supervised GRL baselines, it defines an InfoNCE contrastiveness loss, which contrasts embeddings at the node level from two different augmented views.
    \item SubG-Con \cite{jiao2020sub} Sub-graph Contrast (SubG-Con) defines a self-supervised graph contrastive learning strategy, which extracts regional structure information from the correlation between central nodes and their sampled subgraphs (i.e., a subgraph built by central nodes and its close neighbors.).
\end{itemize}
The experiment result is shown in Table \ref{tab: classification results}. Here we adopt classification accuracy as the evaluation metrics. Every data entry in the table is the averaged result and associated standard deviation over ten individual runs. 
According to this table, we have made the following observations:
\begin{itemize}
    \item On all datasets, our proposed approach G-Zoom has surpassed all baselines and achieved the best performance. Notably, G-Zoom exceeds 
    the best result of its self-supervised GRL counterparts by more than 1\% on Cora and around 1\% on CiteSeer and Coauthor CS. 
    This is mainly because G-Zoom successfully extracts useful clues from different perspectives through contrastive learning via the proposed adjusted zooming scheme, by which G-Zoom can inspect graphs comprehensively and learn effective embeddings.
    \item G-Zoom successfully handles a large-scale dataset (i.e., ogbn-arxiv), and achieves the best performance on the large-scale dataset. At the same time, almost all self-supervised GRL baselines except for SubG-Con, failed to handle it within the limited GPU memory budget. \change{This is because G-Zoom adopts a tailored sub-sampling strategy and a scalable graph diffusion approach with power iteration. Thus, it has the potential to handle large graph datasets.} Though SubG-Con successfully processes ogbn-arxiv, their performance is not satisfying compared with \changesecond{the result of G-Zoom}. This is probably because SubG-Con only contrasts central nodes to their close neighbors, which overlooks the rich global information embodied in a large graph. In contrast, our approach has a more comprehensive contrastive learning scheme.
    \item Without the guidance of labels, G-Zoom is competitive even compared with supervised GNN approaches including GCN, GAT, SGC, and CG3. Our method 
    outperforms all these supervised counterparts on all datasets by large margins, which shows \changesecond{the superiority of G-Zoom} in graph representation learning.
\end{itemize}

    \begin{table}[t]
    \centering
        \caption{Performance comparison of two model variants $\textbf{G-Zoom}_{w/oSample}$ and $\textbf{G-Zoom}_{power}$ with  $\textbf{G-Zoom}$ on three benchmark datasets.}
        \begin{tabular}{lrrr}
        \toprule
        \textbf{Dataset} &  \textbf{Cora} &  \textbf{CiteSeer} &  \textbf{Coauthor CS} \\
        \midrule
         $\textbf{G-Zoom}_{w/oSample}$ &  84.0 &      73.8 &         94.0 \\
         $\textbf{G-Zoom}_{power}$    &  84.2 &      74.4 &         94.4 \\
        \midrule
          $\textbf{G-Zoom}$    &  84.7 &      74.2 &         94.9 \\
        \bottomrule
        \end{tabular}
    \label{tab:ab_sub_diff}
    \end{table}

\begin{table*}[h]
	\footnotesize
	\caption{Performance comparison of using different augmentation strategies on five benchmark datasets. Here ``/'' separates the augmentation techniques used for each view. For example, ``GS/GS+GD'' means using GS to generate the first augmented view, and applying both GS and GD to generate the second augmented view.}
    	\begin{tabular*}{1.0\textwidth}{@{\extracolsep{\fill}}llcccccc}
    		\toprule
    		Types & Augmentation & Cora & Citeseer & Coauthor CS &Coauthor Physics & ogbn-arxiv\\
    		\midrule
                & \change{GS/GS} & \change{83.6 $\pm{0.3}$} & \change{74.0 $\pm{0.5}$}  & \change{93.0 $\pm{0.4}$} & \change{94.7  $\pm{0.4}$} & \change{68.8 $\pm{0.2}$}  \\
                 & GS/GS+GD+ED & 74.1 $\pm{1.3}$ & 73.4 $\pm{0.5}$ & 93.1 $\pm{0.7}$ & 95.7 $\pm{0.1}$ & 69.0 $\pm{0.1}$ \\
               \textbf{Structural}  & GS/GS+GD & \textbf{84.7 $\bm{\pm{0.4}}$} & \textbf{74.2 $\bm{\pm{0.4}}$}  & \textbf{94.9 $\bm{\pm{0.5}}$} & \textbf{96.5 $\bm{\pm{0.4}}$} & \textbf{70.2 $\bm{\pm{0.1}}$}  \\
    		& GS+ED/GS+GD+ED & 74.1 $\pm{1.2}$ & 73.0 $\pm{0.4}$ & 92.8  $\pm{0.6}$ & 95.9 $\pm{0.2}$ & 69.0 $\pm{0.2}$ \\
    		& GS+ED/GS+GD & 84.3 $\pm{0.3}$ & 73.2 $\pm{0.4}$  & 94.9$\pm{0.4}$ & 96.3 $\pm{0.2}$ & 69.1 $\pm{0.4}$\\
    		\midrule
    			& GS/GS+GD+NFD & 84.3 $\pm{0.3}$ & 73.6 $\pm{0.3}$  & 94.2 $\pm{0.2}$ & 96.3 $\pm{0.3}$& 69.7 $\pm{0.1}$ \\
    		    \textbf{Structural+Attributive}& GS+NFD/GS+NFD & 83.5 $\pm{0.3}$ & 74.2 $\pm{0.5}$  & 94.5 $\pm{0.4}$ & 96.0 $\pm{0.6}$ & 69.4 $\pm{0.3}$\\
    			& GS+NFD/GS+GD & 84.1 $\pm{0.5}$ & 74.1 $\pm{0.2}$  & 94.4 $\pm{0.2}$ & 96.0 $\pm{0.3}$ & 69.4 $\pm{0.3}$  \\
    			& GS+NFD/GS+GD+NFD & 84.2 $\pm{0.5}$ & 73.8 $\pm{0.7}$  & 94.2 $\pm{0.6}$ & 95.7 $\pm{0.4}$ & 69.5 $\pm{0.3}$ \\
    		\bottomrule
    	\end{tabular*}
\label{tab: Augmentation}
\end{table*}
\subsection{Ablation Study}
\label{section:V-D}

As is mentioned in section \ref{section:IV-C}, G-Zoom establishes the contrastive paths from three perspectives, 
namely micro, meso and macro, to encode both localized and global information comprehensively. 
To shed light on our proposed adjusted zooming-based contrastiveness, 
we present 
the experiment results of G-Zoom variants which only adopt a specific level 
of loss or employ the combination of two contrastive losses 
on five benchmark datasets, as shown in Table \ref{tab: Ablation Study}. For a better illustration, we use $\textbf{G-Zoom}_{w/o Micro}$, $\textbf{G-Zoom}_{w/o Meso}$ , and $\textbf{G-Zoom}_{w/o Macro}$  to denote G-Zoom variants without considering the micro, meso, and macro perspective, respectively. Also, $\textbf{G-Zoom}_{Micro}$, $\textbf{G-Zoom}_{Meso}$, and $\textbf{G-Zoom}_{Macro}$ represent the reduced models using only one level of contrastiveness. From Table \ref{tab: Ablation Study}, \change{we can observe that $\textbf{G-Zoom}_{w/o Meso}$  and $\textbf{G-Zoom}_{w/o Macro}$  generally surpass variants with only one level of contrastiveness. This indicates that adding $\textbf{G-Zoom}_{w/o Micro}$  to $\textbf{G-Zoom}_{Meso}$  or $\textbf{G-Zoom}_{Macro}$  improves model performance, to which the inclusion of two distinct viewpoints contributes. While the meso or macro viewpoint can provide rich global information, the micro viewpoint  extracts the most fine-grained localized information from node-node comparison.}

\changesecond{Also, we can see that without either the meso or macro contrastiveness, G-Zoom cannot achieve the best performance in 4 out of 5 datasets, which indicates that the combination of these two components is effective.} We conjecture this is because the meso module can force the model to explore the fine-grained information embodied in a specific scale of a graph, which can be easily neglected from a macro-level viewpoint.

\changesecond{To evaluate the impact of the subsampling and the proposed graph diffusion technique (defined in Section \ref{section:IV-E}) on G-Zoom, we compare two model variants $\textbf{G-Zoom}_{w/oSample}$ (i.e., G-Zoom without subsampling) and $\textbf{G-Zoom}_{power}$ (i.e., G-Zoom with power-iteration diffusion) with G-Zoom. The experiment result is reported in Table \ref{tab:ab_sub_diff}. From the table, we observe that without subsampling, the performance of G-Zoom degrades. We conjecture this is because the subsampling creates different subgraphs in each iteration, which increases the difficulty of the self-supervised learning tasks with changing topology of the input graph. Thus, subsampling can improve the model performance. In addition, using the power-iteration based diffusion, the performance of $\textbf{G-Zoom}_{power}$ is comparatively lower than G-Zoom with the original graph diffusion. This is probably because the power-iteration graph diffusion is an approximation of the original diffusion. Thus, it may not be as comprehensive as the original diffusion in global information extraction.}

\subsection{Augmentation}
\label{section:V-E}
To find the most appropriate augmentation scheme for G-Zoom, 
we have conducted experiments 
using eight different graph augmentations, which 
can be categorized into two types: ``Structural'' and ``Structural+Attributive''. Please \changethird{note} that there is no attributive only augmentation scheme in our model because to ensure the scalability of G-Zoom, we have 
to apply GS as a structural augmentation technique to create subgraphs as the inputs of graph encoder. 

The results of the experiment are presented in Table \ref{tab: Augmentation}. As shown in the table, we can see that the most effective augmentation scheme is GS/GS+GD, which has achieved the best performance on all benchmark datasets. Except for its effectiveness, GS/GS+GD is also comparatively more efficient in computation \change{than most augmentation schemes. Without diffusion (i.e., the GS/GS scheme), the model performance degrades on all five datasets, which validates the effectiveness of the diffusion augmentation.} It is interesting to observe that adding ED \changethird{to} the diffused graph degrades the model performance by a large amount. Specifically, 
the worst performance for each dataset is either obtained by GS/GS+GD+ED or GS+ED/GS+GD+ED. We conjecture that this is because ED on diffused graph corrupts the augmented view too much. The graph diffusion operation is extending the receptive field in messaging passing by densifying a graph. As a result, the diffused graph has much more edges than the input graph, and thus applying ED on the diffused graph may corrupt too much of the underlying linkages. 
Also, it is worth noting that integrating NFD in an augmentation scheme negatively affects \changesecond{the performance of G-Zoom}.

\begin{figure*}
     \centering
     \begin{subfigure}[b]{0.32\textwidth}
         \centering
         \includegraphics[width=\textwidth]{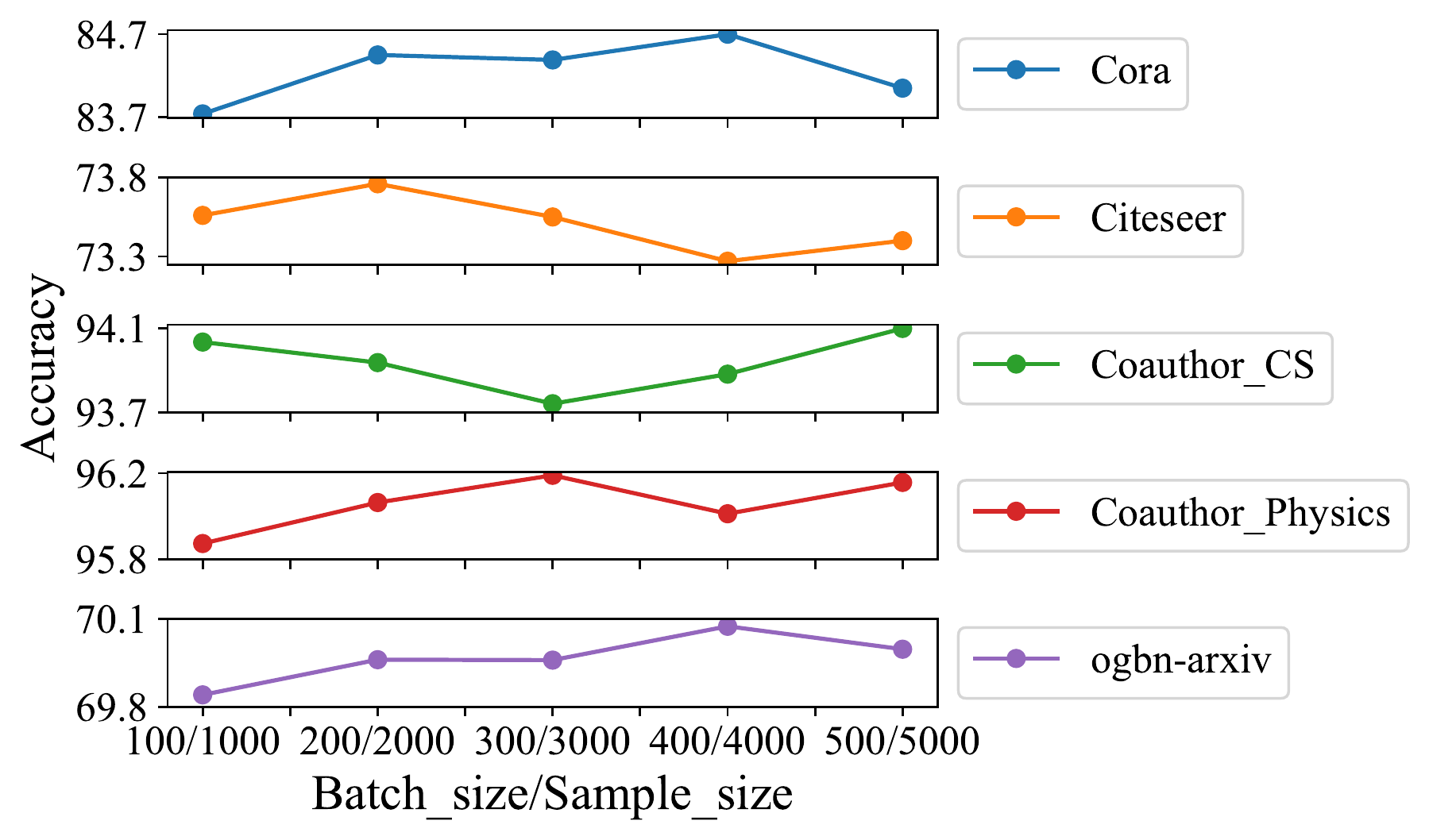}
         \caption{Batch size/Sample size}
         \label{fig:parameter(1)}
     \end{subfigure}
     \hfill
     \begin{subfigure}[b]{0.32\textwidth}
         \centering
         \includegraphics[width=\textwidth]{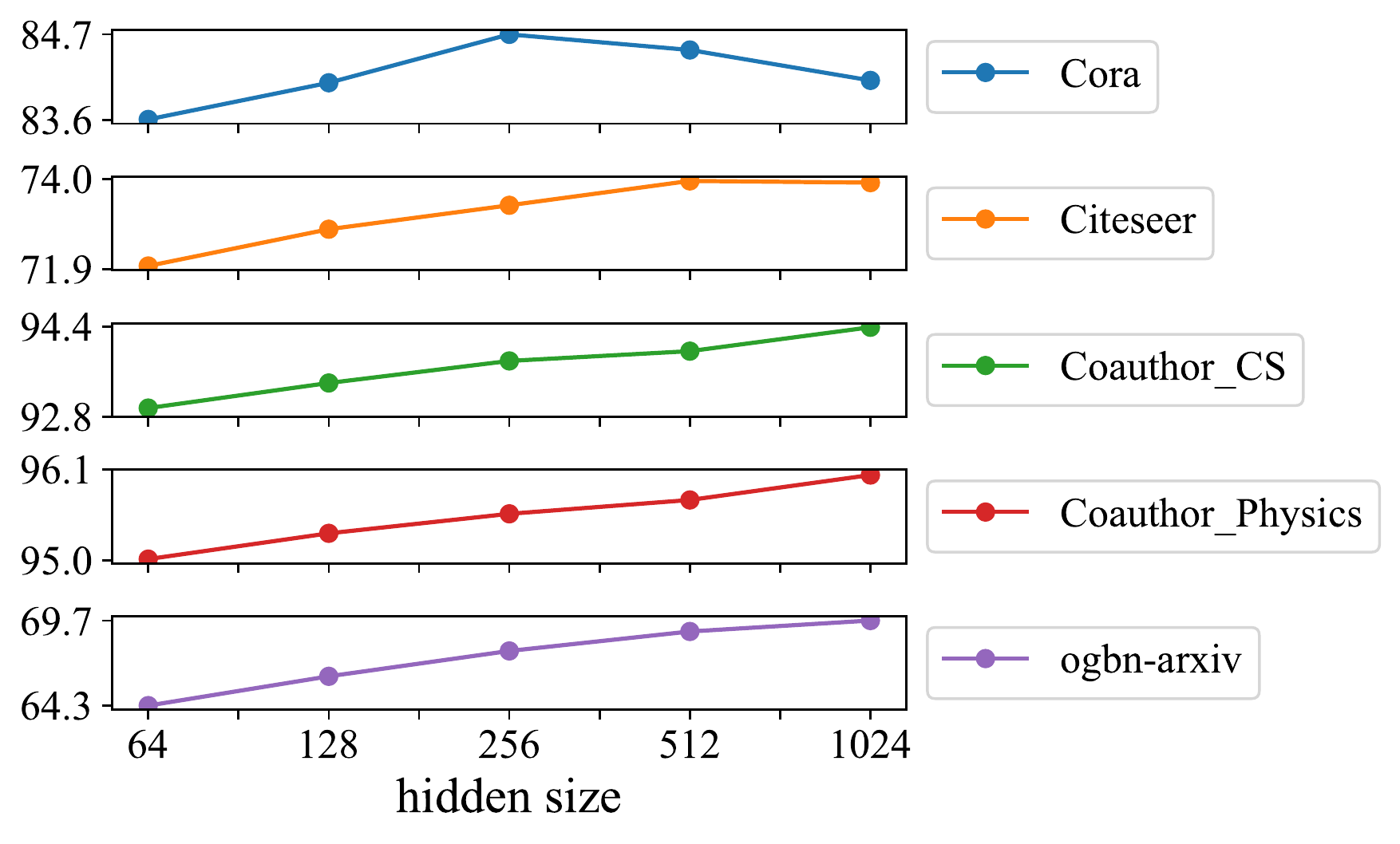}
         \caption{hidden size}
         \label{fig:parameter(2)}
     \end{subfigure}
     \hfill
     \begin{subfigure}[b]{0.32\textwidth}
         \centering
         \includegraphics[width=\textwidth]{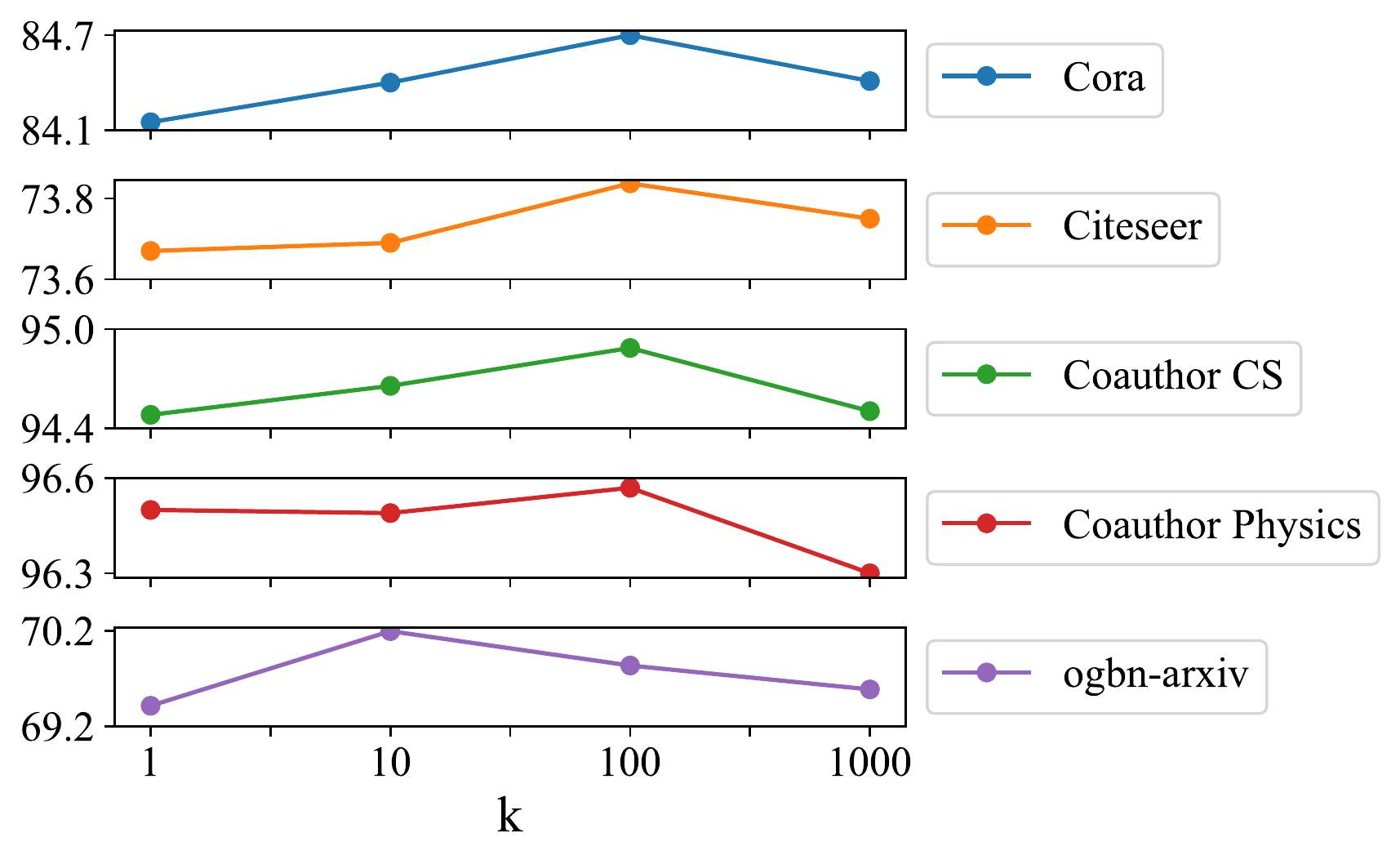}
         \caption{k}
         \label{fig:parameter(3)}
     \end{subfigure}
        \caption{The experiment results for our parameter study. The subfigure(a), (b), (c) represents the influence of parameters including batch size $B$ and sample size $P$, hidden size $h$ and number of top neighbors $k$ to \changesecond{the performance of G-Zoom} on five datasets.}
        \label{fig:three graphs}
\end{figure*}

\begin{figure*}[h]
     \centering
     \begin{subfigure}[b]{0.15\textwidth}
         \centering
         \includegraphics[width=\textwidth]{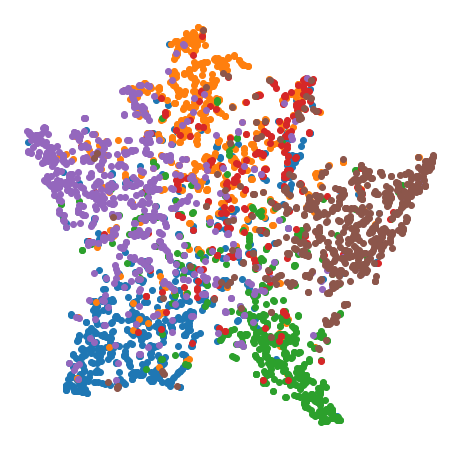}
         \caption{GCN \\ (NMI: 0.38)}
         \label{fig:tsne-cora_1}
     \end{subfigure}
     \hfill
     \begin{subfigure}[b]{0.15\textwidth}
         \centering
         \includegraphics[width=\textwidth]{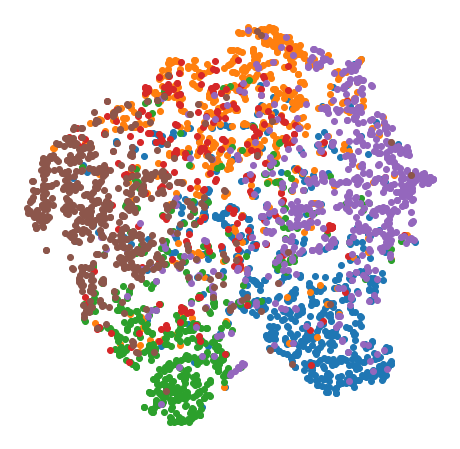}
         \caption{DGI \\ (NMI: 0.39)}
         \label{fig:tsne-dgi_1}
     \end{subfigure}
     \hfill
     \begin{subfigure}[b]{0.15\textwidth}
         \centering
         \includegraphics[width=\textwidth]{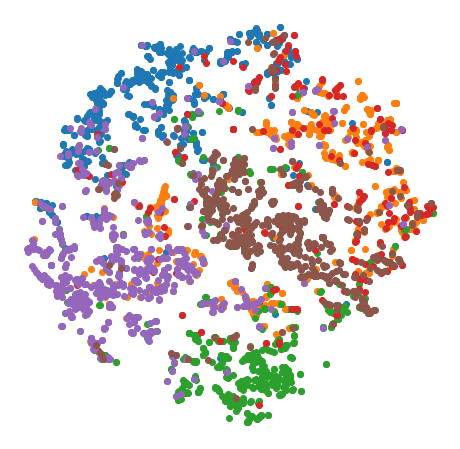}
         \caption{GRACE \\ (NMI: 0.40)}
         \label{fig:tsne-grace_1}
     \end{subfigure}
     \hfill
     \begin{subfigure}[b]{0.15\textwidth}
         \centering
         \includegraphics[width=\textwidth]{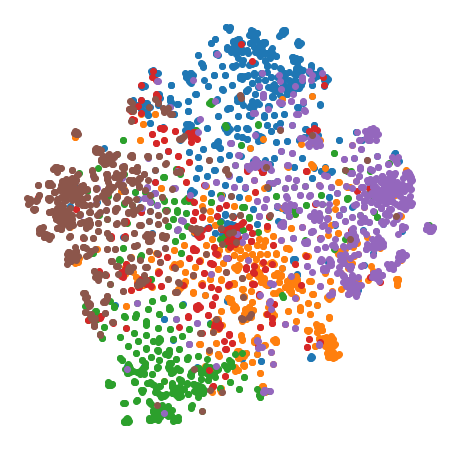}
         \caption{SubG-Con \\ (NMI: 0.40)}
         \label{fig:tsne subg-con_1}
     \end{subfigure}
    \hfill
     \begin{subfigure}[b]{0.15\textwidth}
     \centering
     \includegraphics[width=\textwidth]{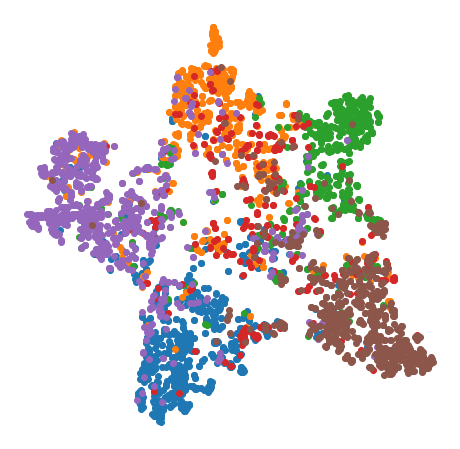}
     \caption{G-Zoom \\ (NMI: 0.46)}
     \label{fig:tsne G-Zoom_1}
     \end{subfigure}
        \caption{t-SNE visualization of G-Zoom and five baselines of the CiteSeer dataset.}
        \label{fig:tsne_1}
\end{figure*}

\begin{table}[h]
	\footnotesize
	\centering
	\caption{Memory consumption of 4 GCL baselines and G-Zoom on five benchmark datasets. The unit for each data entry is MB.}
	{
        \begin{tabular}{llllll}
        \toprule
        \textbf{Method} &  \textbf{Cora} & \textbf{CiteSeer} & \specialcell{\textbf{Coauthor}\\ \textbf{CS}} & \specialcell{\textbf{Coauthor} \\ \textbf{Physics}} & \textbf{ogbn-arxiv} \\
        \midrule
        \textbf{DGI}      &  1,523 &     1,667 &        6,557 &             3,685 &        OOM \\
        \textbf{GMI}      &  3,927 &     7,605 &         OOM &              OOM &        OOM \\
        \textbf{MVGRL}    &  9,695 &     9,867 &       13,859 &            13,777 &        OOM \\
        \textbf{GRACE}    &  1,499 &     1,024 &       19,587 &            12,645 &        OOM \\
        \textbf{SubG-Con} &  1,586 &     1,163 &        2,706 &             8,598 &        601 \\
        \textbf{G-Zoom}   &   759 &      991 &        3,889 &             2,429 &       1,591 \\
        \bottomrule
        \end{tabular}
    }
\label{tab: memory}
\end{table}

\subsection{Memory Efficiency}
To qualitatively compare the memory required by each algorithm, we report the memory consumption of G-Zoom and the baselines on five benchmark datasets. Specifically, we run the baselines based on their default setting, and set the batch size $B$ and the sample size $P$ to 200/2000 for the first two datasets, and 500/5000 for the remaining datasets when running G-Zoom. As shown in Table \ref{tab: memory}, the memory consumption of G-Zoom is the lowest in 3 out of 5 datasets of all the methods, which clearly indicates the memory efficiency of G-Zoom. Another algorithm that can successfully complete all the tasks is SubG-Con, as it employs a small-size subgraph generation approach for graph representation learning. The memory consumption of G-Zoom is mainly controlled by the batch size $B$ and the sample size $P$. With lower $B$ and $P$, the memory consumption of G-Zoom can be even lower, and thus it can handle large datasets.\\

\subsection{Parameter Study}
\label{section:V-G}
In this subsection, we investigate the influences of four key parameters on the performance of our proposed model, 
which are batch size $B$, sample size $P$, hidden size $d'$ and parameter $k$ for choosing the number of closest 
neighbors for each target node on five benchmark datasets.

\subsubsection{Batch size $B$ and sample size $P$}
In this experiment, we change the value of
$B$ and $P$ simultaneously to see how these two parameters affect 
\changesecond{the performance of G-Zoom} and the experiment results are reported in Figure \ref{fig:parameter(1)}.
As illustrated in subsection \ref{section:IV-A}, increasing $B$ requires larger $P$ since the subgraph with size $P$ needs to contain 
$B$ nodes and their top-$k$ neighbors. 
Therefore, we choose five $B$ values from 100 to 500, incremented by 100, and their corresponding $P$ from 1000 to 5000, incremented by 1000.
As shown in Figure \ref{fig:parameter(1)}, changing $B$ and $P$ have different impacts on model performance on other datasets. 
We can observe that compared with small graphs such as Cora and Citesser, adopting large $B$ and $P$ on bigger datasets such as ogbn-arxiv, coauthor CS and Physics in G-Zoom achieve improved results. This is probably because increasing $B$ and $P$ can help better encode global information in large graphs, while a small size $P$ is enough to contain a large proportion of nodes in small datasets. Thus, choosing appropriate $B$ and $P$ for G-Zoom is based on the size of datasets to be trained on.

\subsubsection{Hidden size $d'$}
We investigate the sensitivity of the hidden size $d'$ for our proposed framework G-Zoom. Here we choose 64, 128, 256, 512, and 1024 as the hidden size of our model 
to be explored to see how increasing $d'$ impacts the accuracy of G-Zoom on benchmark datasets. 
From Figure \ref{fig:parameter(2)}, there is an obvious upward 
trend on 
all datasets 
except for Cora, which indicates $d'$ has a positive relationship with the performance of G-Zoom. The possible reason is that rising $d'$ will increase the number of trainable parameters, which extends the expressive ability of our model. 
However, without sufficient data, 
too many parameters may cause the over-fitting problem, which degrades \changesecond{ the performance of a model~\cite{hawkins2004problem}}. This may explain why with a large $d'$, G-Zoom suffers from lower performance on Cora since it contains a limited number of nodes and edges. 

\subsubsection{Neighborhood sampling size $k$}
To explore the significance of $k$ on different datasets, we conducted experiments by using four different $k$ values, which are 1, 10, 100, and 1000. 
Our results are 
demonstrated in Figure \ref{fig:parameter(3)}, where we have a similar trend over all datasets: The expressiveness of the model has been boosted by increasing $k$ from 1 to 10 or 100. However, a large number of sampled neighbors (e.g., more than 100) hurts the performance. 

\subsection{Visualization}
\label{section:V-H}
For an intuitive illustration, we visualize the learned embeddings of four baselines and our method (i.e., GCN, DGI, GRACE, SubG-Con, and G-Zoom)  \change{on CiteSeer} via using the t-SNE method \cite{van2008visualizing}. The visualized representation of these embeddings is given in Figure \ref{fig:tsne_1}. As can be observed, the 2D projection of embeddings generated by G-Zoom exhibits more discernible clusters than embeddings learned by other approaches. Qualitatively, the visualization demonstrates the effectiveness of G-Zoom in graph self-supervised learning. \change{To verify this, we further added  a new experiment, which applie\changethird{d} K-mean clustering to the learned representations of the five GCL methods on CiteSeer and used the normalized mutual information (NMI) as the metric to evaluate the clustering goodness. The experiment results are shown in the caption of each subfigure in Figure \ref{fig:tsne_1}. From the results, it can be seen that the learned representations generated by G-Zoom achieved the highest NMI score, indicating better performance.}

\section{Conclusion}
In this paper, we have proposed a novel framework G-Zoom, a graph self-supervised learning approach powered by contrastive adjusted zooming. In our framework, G-Zoom consists of two components: 
augmented graph encoding 
and adjusted zooming powered contrastive learning. 
While the former component can generate multiple augmented views, the latter one facilitates the construction of underlying contrastive paths 
from three viewpoints: 
micro, meso, and macro. G-Zoom can effectively extract 
both local and global self-supervision 
signals of a graph 
by progressively inspecting it 
from \changethird{the} finest details to the coarse landscape. Extensive experiments have demonstrated the effectiveness and superiority of our model over existing self-supervised, as well as some supervised GRL baselines by large margins. In the future, we plan to exploit automatic machine learning techniques \cite{zhang2020one} to search for the best neural architecture for graph representation learning. 

\bibliographystyle{IEEEtran}
\bibliography{ZOOM}

\begin{thebibliography}{10}
\providecommand{\url}[1]{#1}
\csname url@samestyle\endcsname
\providecommand{\newblock}{\relax}
\providecommand{\bibinfo}[2]{#2}
\providecommand{\BIBentrySTDinterwordspacing}{\spaceskip=0pt\relax}
\providecommand{\BIBentryALTinterwordstretchfactor}{4}
\providecommand{\BIBentryALTinterwordspacing}{\spaceskip=\fontdimen2\font plus
\BIBentryALTinterwordstretchfactor\fontdimen3\font minus
  \fontdimen4\font\relax}
\providecommand{\BIBforeignlanguage}[2]{{%
\expandafter\ifx\csname l@#1\endcsname\relax
\typeout{** WARNING: IEEEtran.bst: No hyphenation pattern has been}%
\typeout{** loaded for the language `#1'. Using the pattern for}%
\typeout{** the default language instead.}%
\else
\language=\csname l@#1\endcsname
\fi
#2}}
\providecommand{\BIBdecl}{\relax}
\BIBdecl

\bibitem{kipf2016semi}
T.~N. Kipf and M.~Welling, ``Semi-supervised classification with graph
  convolutional networks,'' \emph{ICLR}, 2017.

\bibitem{zhang2022trustworthy}
H.~Zhang, B.~Wu, X.~Yuan, S.~Pan, H.~Tong, and J.~Pei, ``Trustworthy graph
  neural networks: Aspects, methods and trends,'' \emph{arXiv preprint
  arXiv:2205.07424}, 2022.

\bibitem{sun2019infograph}
F.-Y. Sun, J.~Hoffmann, V.~Verma, and J.~Tang, ``Infograph: Unsupervised and
  semi-supervised graph-level representation learning via mutual information
  maximization,'' \emph{ICLR}, 2020.

\bibitem{schutt2017schnet}
K.~T. Sch{\"u}tt, P.-J. Kindermans, H.~E. Sauceda, S.~Chmiela, A.~Tkatchenko,
  and K.-R. M{\"u}ller, ``Schnet: A continuous-filter convolutional neural
  network for modeling quantum interactions,'' \emph{NIPS}, 2017.

\bibitem{xia2021graph}
F.~Xia, K.~Sun, S.~Yu, A.~Aziz, L.~Wan, S.~Pan, and H.~Liu, ``Graph learning: A
  survey,'' \emph{IEEE Transactions on Artificial Intelligence}, 2021.

\bibitem{wan2021dual}
S.~Wan, S.~Pan, P.~Zhong, X.~Chang, J.~Yang, and C.~Gong, ``Dual interactive
  graph convolutional networks for hyperspectral image classification,''
  \emph{IEEE Transactions on Geoscience and Remote Sensing}, 2021.

\bibitem{liu2021anomaly}
Y.~Liu, Z.~Li, S.~Pan, C.~Gong, C.~Zhou, and G.~Karypis, ``Anomaly detection on
  attributed networks via contrastive self-supervised learning,'' \emph{IEEE
  Transactions on Neural Networks and Learning Systems}, 2021.

\bibitem{yizhen2021hetergraph}
Y.~Zheng, C.~L. Vincent, Z.~Wu, and S.~Pan, ``Heterogeneous graph attention
  network for small and medium-sized enterprises bankruptcy prediction,''
  \emph{PAKDD}, 2021.

\bibitem{dai2016discriminative}
H.~Dai, B.~Dai, and L.~Song, ``Discriminative embeddings of latent variable
  models for structured data,'' in \emph{ICML}.\hskip 1em plus 0.5em minus
  0.4em\relax PMLR, 2016.

\bibitem{velivckovic2017graph}
P.~Veli{\v{c}}kovi{\'c}, G.~Cucurull, A.~Casanova, A.~Romero, P.~Lio, and
  Y.~Bengio, ``Graph attention networks,'' \emph{ICLR}, 2018.

\bibitem{wu2019simplifying}
F.~Wu, A.~Souza, T.~Zhang, C.~Fifty, T.~Yu, and K.~Weinberger, ``Simplifying
  graph convolutional networks,'' in \emph{ICML}.\hskip 1em plus 0.5em minus
  0.4em\relax PMLR, 2019.

\bibitem{velickovic2019deep}
P.~Velickovic, W.~Fedus, W.~L. Hamilton, P.~Li{\`o}, Y.~Bengio, and R.~D.
  Hjelm, ``Deep graph infomax.'' in \emph{ICLR}, 2019.

\bibitem{zhu2020deep}
Y.~Zhu, Y.~Xu, F.~Yu, Q.~Liu, S.~Wu, and L.~Wang, ``Deep graph contrastive
  representation learning,'' \emph{ICML}, 2020.

\bibitem{peng2020graph}
Z.~Peng, W.~Huang, M.~Luo, Q.~Zheng, Y.~Rong, T.~Xu, and J.~Huang, ``Graph
  representation learning via graphical mutual information maximization,'' in
  \emph{WWW}, 2020.

\bibitem{jiao2020sub}
Y.~Jiao, Y.~Xiong, J.~Zhang, Y.~Zhang, T.~Zhang, and Y.~Zhu, ``Sub-graph
  contrast for scalable self-supervised graph representation learning,''
  \emph{ICDM}, 2020.

\bibitem{liu2021graph}
Y.~Liu, S.~Pan, M.~Jin, C.~Zhou, F.~Xia, and P.~S. Yu, ``Graph self-supervised
  learning: A survey,'' \emph{arXiv preprint arXiv:2103.00111}, 2021.

\bibitem{lecun2015deep}
Y.~LeCun, Y.~Bengio, and G.~Hinton, ``Deep learning,'' \emph{nature}, 2015.

\bibitem{bruna2013spectral}
J.~Bruna, W.~Zaremba, A.~Szlam, and Y.~LeCun, ``Spectral networks and locally
  connected networks on graphs,'' \emph{ICLR}, 2014.

\bibitem{defferrard2016convolutional}
M.~Defferrard, X.~Bresson, and P.~Vandergheynst, ``Convolutional neural
  networks on graphs with fast localized spectral filtering,'' \emph{NIPS},
  2016.

\bibitem{henaff2015deep}
M.~Henaff, J.~Bruna, and Y.~LeCun, ``Deep convolutional networks on
  graph-structured data,'' \emph{arXiv preprint arXiv:1506.05163}, 2015.

\bibitem{shuman2013emerging}
D.~I. Shuman, S.~K. Narang, P.~Frossard, A.~Ortega, and P.~Vandergheynst, ``The
  emerging field of signal processing on graphs: Extending high-dimensional
  data analysis to networks and other irregular domains,'' \emph{IEEE signal
  processing magazine}, vol.~30, no.~3, pp. 83--98, 2013.

\bibitem{vaswani2017attention}
A.~Vaswani, N.~Shazeer, N.~Parmar, J.~Uszkoreit, L.~Jones, A.~N. Gomez,
  L.~Kaiser, and I.~Polosukhin, ``Attention is all you need,'' \emph{NIPS},
  2017.

\bibitem{bojchevski2020scaling}
A.~Bojchevski, J.~Klicpera, B.~Perozzi, A.~Kapoor, M.~Blais,
  B.~R{\'o}zemberczki, M.~Lukasik, and S.~G{\"u}nnemann, ``Scaling graph neural
  networks with approximate pagerank,'' in \emph{SIGKDD}, 2020.

\bibitem{zeng2019graphsaint}
H.~Zeng, H.~Zhou, A.~Srivastava, R.~Kannan, and V.~Prasanna, ``Graphsaint:
  Graph sampling based inductive learning method,'' \emph{ICLR}, 2020.

\bibitem{klicpera2019diffusion}
J.~Klicpera, S.~Wei{\ss}enberger, and S.~G{\"u}nnemann, ``Diffusion improves
  graph learning,'' \emph{NIPS}, 2019.

\bibitem{tang2020cgd}
C.~Tang, X.~Liu, X.~Zhu, E.~Zhu, Z.~Luo, L.~Wang, and W.~Gao, ``Cgd: Multi-view
  clustering via cross-view graph diffusion,'' in \emph{AAAI}, 2020.

\bibitem{liao2019lanczosnet}
R.~Liao, Z.~Zhao, R.~Urtasun, and R.~S. Zemel, ``Lanczosnet: Multi-scale deep
  graph convolutional networks,'' \emph{ICLR}, 2019.

\bibitem{wu2021learning}
M.~Wu, S.~Pan, L.~Du, and X.~Zhu, ``Learning graph neural networks with
  positive and unlabeled nodes,'' \emph{ACM Transactions on Knowledge Discovery
  from Data}, 2021.

\bibitem{ji2021survey}
S.~Ji, S.~Pan, E.~Cambria, P.~Marttinen, and S.~Y. Philip, ``A survey on
  knowledge graphs: Representation, acquisition, and applications,'' \emph{IEEE
  Transactions on Neural Networks and Learning Systems}, 2021.

\bibitem{zhang2020relational}
Z.~Zhang, F.~Zhuang, H.~Zhu, Z.~Shi, H.~Xiong, and Q.~He, ``Relational graph
  neural network with hierarchical attention for knowledge graph completion,''
  in \emph{AAAI}, 2020.

\bibitem{zhao2019t}
L.~Zhao, Y.~Song, C.~Zhang, Y.~Liu, P.~Wang, T.~Lin, M.~Deng, and H.~Li,
  ``T-gcn: A temporal graph convolutional network for traffic prediction,''
  \emph{IEEE Transactions on Intelligent Transportation Systems}, 2019.

\bibitem{kipf2016variational}
T.~N. Kipf and M.~Welling, ``Variational graph auto-encoders,'' \emph{NIPS},
  2016.

\bibitem{pan2019learning}
S.~Pan, R.~Hu, S.-f. Fung, G.~Long, J.~Jiang, and C.~Zhang, ``Learning graph
  embedding with adversarial training methods,'' \emph{IEEE Transactions on
  Cybernetics}, vol.~50, no.~6, pp. 2475--2487, 2019.

\bibitem{grover2016node2vec}
A.~Grover and J.~Leskovec, ``node2vec: Scalable feature learning for
  networks,'' in \emph{SIGKDD}, 2016.

\bibitem{perozzi2014deepwalk}
B.~Perozzi, R.~Al-Rfou, and S.~Skiena, ``Deepwalk: Online learning of social
  representations,'' in \emph{SIGKDD}, 2014.

\bibitem{hamilton2017inductive}
W.~L. Hamilton, R.~Ying, and J.~Leskovec, ``Inductive representation learning
  on large graphs,'' \emph{NIPS}, 2017.

\bibitem{hassani2020contrastive}
K.~Hassani and A.~H. Khasahmadi, ``Contrastive multi-view representation
  learning on graphs,'' in \emph{ICML}.\hskip 1em plus 0.5em minus 0.4em\relax
  PMLR, 2020.

\bibitem{chen2020simple}
T.~Chen, S.~Kornblith, M.~Norouzi, and G.~Hinton, ``A simple framework for
  contrastive learning of visual representations,'' in \emph{ICML}.\hskip 1em
  plus 0.5em minus 0.4em\relax PMLR, 2020.

\bibitem{hjelm2018learning}
R.~D. Hjelm, A.~Fedorov, S.~Lavoie-Marchildon, K.~Grewal, P.~Bachman,
  A.~Trischler, and Y.~Bengio, ``Learning deep representations by mutual
  information estimation and maximization,'' \emph{ICLR}, 2019.

\bibitem{caron2020unsupervised}
M.~Caron, I.~Misra, J.~Mairal, P.~Goyal, P.~Bojanowski, and A.~Joulin,
  ``Unsupervised learning of visual features by contrasting cluster
  assignments,'' \emph{NIPS}, 2020.

\bibitem{he2020momentum}
K.~He, H.~Fan, Y.~Wu, S.~Xie, and R.~Girshick, ``Momentum contrast for
  unsupervised visual representation learning,'' in \emph{CVPR}, 2020.

\bibitem{grill2020bootstrap}
J.-B. Grill, F.~Strub, F.~Altch{\'e}, C.~Tallec, P.~H. Richemond,
  E.~Buchatskaya, C.~Doersch, B.~A. Pires, Z.~D. Guo, M.~G. Azar \emph{et~al.},
  ``Bootstrap your own latent: A new approach to self-supervised learning,''
  \emph{NIPS}, 2020.

\bibitem{jin2021merit}
M.~Jin, Y.~Zheng, Y.-F. Li, C.~Gong, C.~Zhou, and S.~Pan, ``Multi-scale
  contrastive siamese networks for self-supervised graph representation
  learning,'' \emph{IJCAI}, 2021.

\bibitem{hu2020open}
W.~Hu, M.~Fey, M.~Zitnik, Y.~Dong, H.~Ren, B.~Liu, M.~Catasta, and J.~Leskovec,
  ``Open graph benchmark: Datasets for machine learning on graphs,''
  \emph{arXiv preprint arXiv:2005.00687}, 2020.

\bibitem{oord2018representation}
A.~v.~d. Oord, Y.~Li, and O.~Vinyals, ``Representation learning with
  contrastive predictive coding,'' \emph{arXiv preprint arXiv:1807.03748},
  2018.

\bibitem{paszke2019pytorch}
A.~Paszke, S.~Gross, F.~Massa, A.~Lerer, J.~Bradbury, G.~Chanan, T.~Killeen,
  Z.~Lin, N.~Gimelshein, L.~Antiga \emph{et~al.}, ``Pytorch: An imperative
  style, high-performance deep learning library,'' \emph{NIPS}, 2019.

\bibitem{wan2020contrastive}
S.~Wan, S.~Pan, J.~Yang, and C.~Gong, ``Contrastive and generative graph
  convolutional networks for graph-based semi-supervised learning,'' in
  \emph{AAAI}, 2021.

\bibitem{hawkins2004problem}
D.~M. Hawkins, ``The problem of overfitting,'' \emph{Journal of chemical
  information and computer sciences}, vol.~44, no.~1, pp. 1--12, 2004.

\bibitem{van2008visualizing}
L.~Van~der Maaten and G.~Hinton, ``Visualizing data using t-sne.''
  \emph{Journal of machine learning research}, vol.~9, no.~11, 2008.

\bibitem{zhang2020one}
M.~Zhang, H.~Li, S.~Pan, X.~Chang, C.~Zhou, Z.~Ge, and S.~W. Su, ``One-shot
  neural architecture search: Maximising diversity to overcome catastrophic
  forgetting,'' \emph{IEEE Transactions on Pattern Analysis and Machine
  Intelligence}, 2020.

\end{thebibliography}

\begin{appendices}
\section{Theoretical Analysis of THE Proposed Sampler}
\change{To prove the effectiveness of the proposed sampler, we present the theoretical analysis as follows:}

\begin{figure}[htbp]
    \centering
    \includegraphics[scale = 0.35]{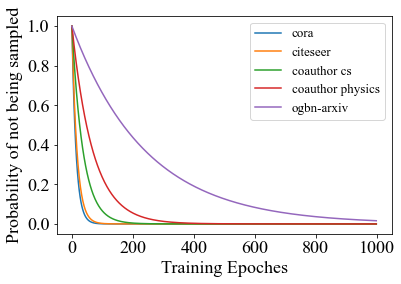}
    \caption{Probability of a node being sampled when training epochs increase for five benchmark datasets. For each dataset, we select their batch size $B$ based on their best setting. The training epoch $E$ is 1000 for all datasets.}
    \label{fig:my_label}
\end{figure}

\change{
\begin{theorem}
Given a graph $\mathcal{G} = (\textbf{X} \in \mathbb{R}^{n \times d},\textbf{A} \in \mathbb{R}^{n \times n})$, the number of nodes in a sampled batch $B$, the number of training epochs $E$, the probability of a node $i$ being sampled from $\mathcal{G}$ is $M$, if $E$ become infinite, we have:
\begin{equation}
    P(i) = \lim_{E \to \infty}(1 - \frac{1}{M})^{BE} = 0,
\end{equation}
where $P(i)$ is the probability of a node $i$ not being sampled after $E$ epochs.
\end{theorem}
}
\change{According to Theorem 1, we can see when $E$ becomes larger, the probability of a node $i$ not being sampled will gradually tend toward zero. As such, we can derive that with a large $E$, our sampler can guarantee all nodes can be sampled, which means the important features of the initial graph will not be lost during training. To validate this, we conduct an experiment to show how $P(i)$ changes when training epochs $E$ increase for five benchmark datasets, as shown in Figure \ref{fig:my_label}, where we can see the probability of a node not being sampled quickly converged to 0 when $E$ increases. Also, the smaller the size of the dataset, the faster the convergence proceeds.}

\end{appendices}
%








\end{document}